\documentclass[10pt,twocolumn,letterpaper]{article}

\usepackage{cvpr}      
\usepackage{graphicx}
\usepackage{booktabs}

\usepackage{lipsum}
\usepackage{multirow}
\usepackage{colortbl}
\usepackage{amsmath}
\usepackage{caption}
\usepackage{verbatim}  
\usepackage{caption}
\usepackage{makecell}
\usepackage{pifont}
\newcommand{\cmark}{\ding{51}} 
\newcommand{\xmark}{\ding{55}} 
\usepackage[breakable]{tcolorbox}
\newtcolorbox{promptbox}[1][]{
  breakable,
  colback=blue!3,
  colframe=blue!70,
  boxrule=0.4pt,
  arc=2pt,
  left=4pt,
  right=4pt,
  top=4pt,
  bottom=4pt,
  fontupper=\ttfamily\small,
  title=#1
}

\definecolor{cvprblue}{rgb}{0.21,0.49,0.74}
\usepackage[pagebackref,breaklinks,colorlinks,allcolors=cvprblue]{hyperref}

\def\paperID{*****} 
\def\confName{CVPR}
\def\confYear{2026}

\title{MegaStyle: Constructing Diverse and Scalable Style Dataset via Consistent Text-to-Image Style Mapping}

\author{Junyao Gao$^{1,2}$\textsuperscript{*} \quad Sibo Liu$^{2}$\textsuperscript{*} \quad Jiaxing Li$^{3}$ \quad Yanan Sun$^{4}$ \quad Yuanpeng Tu$^{6}$ \\ Fei Shen$^{7}$ \quad  Weidong Zhang$^{2}$ \quad Cairong Zhao$^{1,5\dagger}$ \quad Jun Zhang$^{2\dagger}$\\
	$^{1}$Tongji Univeristy, $^{2}$Tencent, $^{3}$Nanyang Technological University, \\ $^{4}$Hong Kong University of Science and Technology, $^{5}$Fuzhou University, \\$^{6}$University of Hong Kong, $^{7}$National University of Singapore\\
	}

\begin{document}
\twocolumn[{%
\renewcommand\twocolumn[1][]{#1}
\maketitle
\vspace{-3em}
\begin{center}
    \centering
    \captionsetup{type=figure}
    \includegraphics[width=\textwidth]{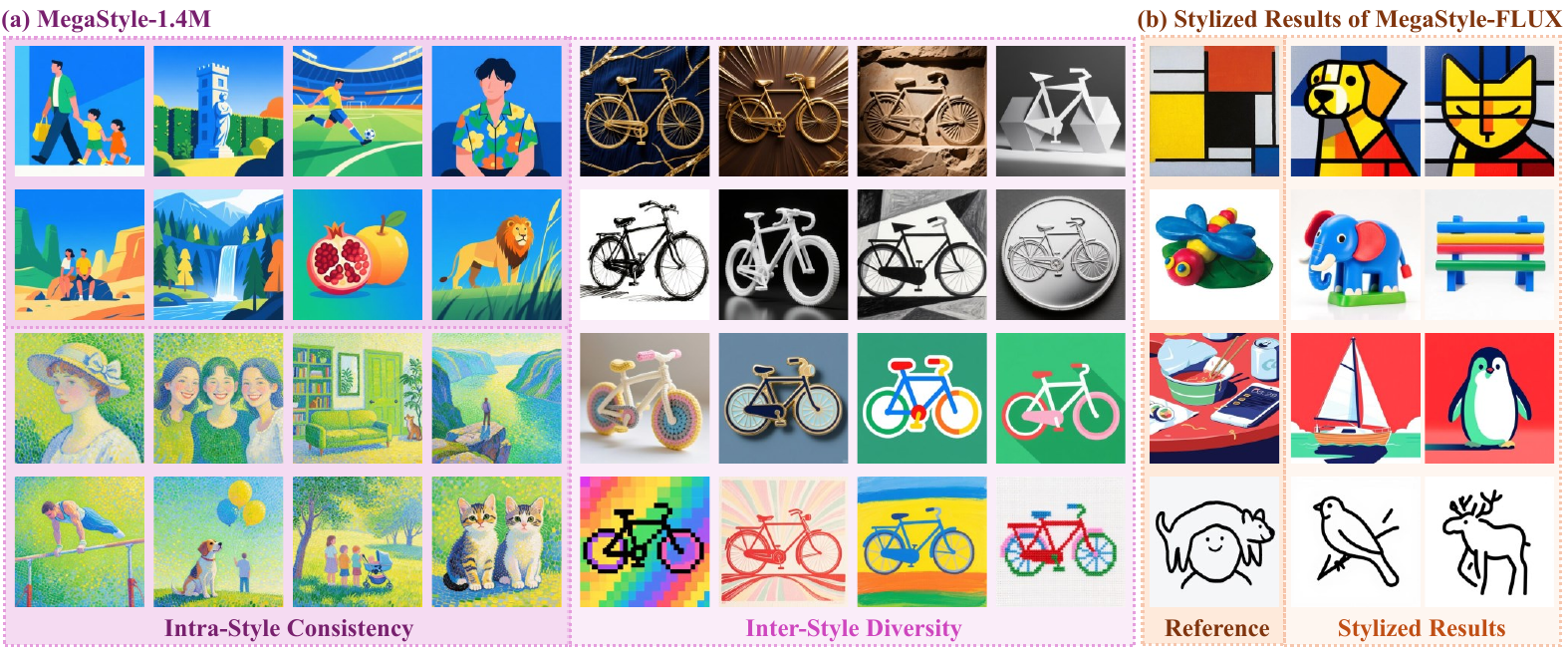}
    \vspace{-2em}
\captionof{figure}{Visualizations of our style dataset (a)MegaStyle-1.4M and the stylized results produced by our style transfer model (b)MegaStyle-FLUX. MegaStyle-1.4M contains style pairs that share the same style but have different content (intra-style consistency), as well as a large number of diverse styles (inter-style diversity). Trained on MegaStyle-1.4M, MegaStyle-FLUX effectively captures nuances—such as color, light, texture and brushwork—across various styles.}
\label{fig:teaser}
\vspace{-0.3cm}
\end{center}}]

\begingroup
\renewcommand\thefootnote{}
\footnote{Work done during Junyao Gao's internship at AIPD, Tencent. \textsuperscript{\ddag}Corresponding authors. *Equal contributions.}
\addtocounter{footnote}{-1}
\endgroup

\begin{abstract}
In this paper, we introduce MegaStyle, a novel and scalable data curation pipeline that constructs an intra-style consistent, inter-style diverse and high-quality style dataset.
We achieve this by leveraging the consistent text-to-image style mapping capability of current large generative models, which can generate images in the same style from a given style description.
Building on this foundation, we curate a diverse and balanced prompt gallery with 170K style prompts and 400K content prompts, and generate a large-scale style dataset MegaStyle-1.4M via content–style prompt combinations.
With MegaStyle-1.4M, we propose style-supervised contrastive learning to fine-tune a style encoder MegaStyle-Encoder for extracting expressive, style-specific representations, and we also train a FLUX-based style transfer model MegaStyle-FLUX.
Extensive experiments demonstrate the importance of maintaining intra-style consistency, inter-style diversity and high-quality for style dataset, as well as the effectiveness of the proposed MegaStyle-1.4M.
Moreover, when trained on MegaStyle-1.4M, MegaStyle-Encoder and MegaStyle-FLUX provide reliable style similarity measurement and generalizable style transfer, making a significant contribution to the style transfer community.
More results are available at our project website \url{https://jeoyal.github.io/MegaStyle/}.
\end{abstract}    
\section{Introduction}
Image style transfer aims to generate stylized images that follow the style of a reference style image and the content provided by the user.
With significant advances in diffusion models \cite{ho2020denoising, nichol2021improved, nichol2021glide,ramesh2022hierarchical, rombach2022high, peebles2023scalable, tu2025playerone, tu2025videoanydoor, tumanyan2023plug, gao2025faceshot, gao2025charactershot}, style transfer has achieved impressive performance \cite{11165480, qi2024deadiff, sohn2024styledrop,zhou2025attention,hertz2023style} and has been widely used in everyday applications such as camera filters and artistic creation.

\begin{figure}
    \centering
    \includegraphics[width=1.0\linewidth]{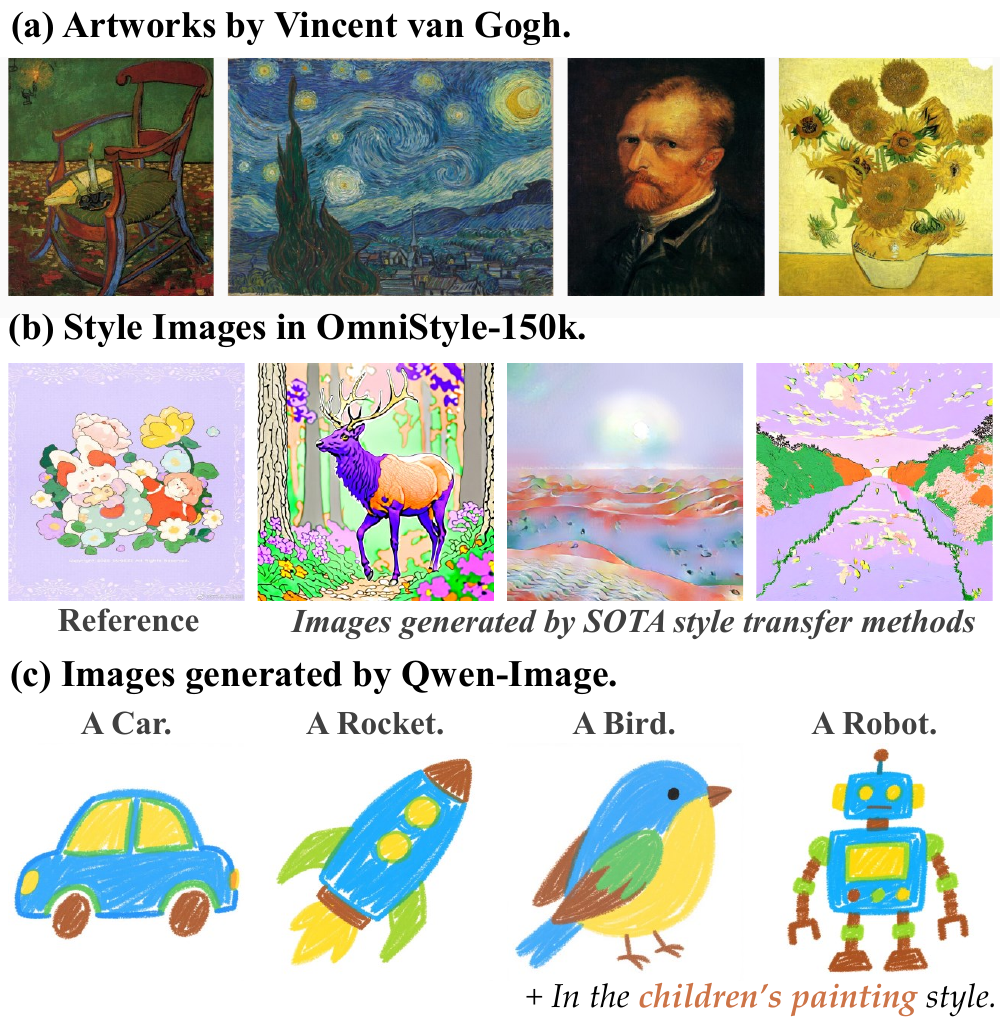}
    \vspace{-6mm}
    \caption{Illustrations of (a) artworks by \textit{Vincent van Gogh}; (b) style images in OmniStyle-150K generated by SOTA style transfer methods \cite{chung2024style, 11165480, xing2024csgo, an2021artflow, hong2023aespa, zhang2022domain} from a reference style image; and (c) images generated by Qwen-Image using the same style description.}
    \vspace{-6mm}
    \label{fig:intro}
\end{figure}

Previous style transfer methods either memorize style from a few reference images into trainable embeddings \cite{zhang2023inversion, gal2022image} or adapters \cite{sohn2024styledrop, hu2021lora}, or use a CLIP \cite{radford2021learning} image encoder to extract style features and inject them as an extra condition to generate stylized images \cite{11165480,liu2023stylecrafter}.
These methods follow a self-supervised training paradigm in which the training target and the reference style image are the same, making it difficult to disentangle style from the tightly coupled image or feature space and leading to content leakage and poor stylized results \cite{liu2023stylecrafter,wang2023styleadapter}. 
A simple yet effective solution is to employ paired supervision—a data-driven training paradigm that has been widely validated in other generative tasks such as editing \cite{vace, sheynin2024emu}—to implicitly model the style transformation using high-quality, diverse style pairs that share the same style but differ in content.
However, style is inherently multi-dimensional and highly discriminative; even minor changes can lead to perceptually different styles during creation. 
As shown in Figure \ref{fig:intro}(a), artworks by \textit{Vincent van Gogh} from the same period can exhibit noticeably different styles.
This makes it difficult to collect style pairs from the Internet.
Additionally, the lack of reliable style similarity measurement \cite{11165480,sohn2024styledrop,liu2023stylecrafter} also hinders the automatic scaling of style datasets.

To address these, IMAGStyle \cite{xing2024csgo} and OmniStyle-150K \cite{wang2025omnistyle} employ state-of-the-art (SOTA) style transfer methods \cite{chung2024style, 11165480, xing2024csgo, an2021artflow, hong2023aespa, zhang2022domain, hu2021lora} to synthesize stylized images from a given reference image.
Yet the inter-style diversity, intra-style consistency, and quality of style pairs in these datasets are heavily constrained by the unstable performance of SOTA style transfer methods.
Specifically, as shown in Figure \ref{fig:intro}(b), the generated images mainly transfer only the basic colors of the reference image, which results in a limited style space.
Beyond color, the texture and brushwork also vary across these images (from left to right: digital illustration, heavy watercolor wash, and flat shading), resulting in inconsistent styles within the style pairs.
Moreover, the generated images exhibit visible artifacts such as color bleeding, haloing, and broken contours.

In this paper, we propose \textbf{MegaStyle}, a scalable data curation pipeline for constructing an intra-style consistent, inter-style diverse and high-quality style dataset.
MegaStyle begins with the observation that SOTA text-to-image (T2I) generative models, such as Qwen-Image \cite{wu2025qwen}, can produce precise, fine-grained responses to textual inputs, which is sufficient for establishing a consistent mapping from a style prompt to a specific image style.
As shown in Figure \ref{fig:intro}(c), with the same style prompt, Qwen-Image generates high-quality style pairs with a consistent style across different contents.
Based on this consistent T2I style mapping, we use vision–language models (VLMs) to caption images from content/style image pools and carefully curate a diverse, balanced prompt gallery comprising 400K content prompts and 170K style prompts.
We then pair each style prompt with numerous content prompts and employ Qwen-Image to generate stylized images from these content–style prompt combinations, forming a large-scale style dataset, MegaStyle-1.4M.
With MegaStyle-1.4M, we propose \textbf{style-supervised contrastive learning} (SSCL) to fine-tune a style encoder named \textbf{MegaStyle-Encoder}, providing style-specific representations for reliable style similarity measurement. 
We also apply the paired supervision to train a Diffusion Transformer (DiT) \cite{peebles2023scalable}-based model FLUX \cite{flux2024}, resulting in \textbf{MegaStyle-FLUX}, which supports generalizable and stable style transfer.

Extensive qualitative and quantitative evaluations demonstrate that MegaStyle-Encoder and MegaStyle-FLUX provide reliable style similarity measurement and generalizable style transfer, outperforming existing baseline methods.
Moreover, ablation studies confirm the effectiveness and advantages of our framework, offering valuable insights to the style transfer community.
The contributions of this paper are summarized as follows:
\begin{itemize}
    \item We propose MegaStyle, a novel and scalable data curation pipeline that first explores consistent T2I style mapping ability from current large generative models to construct intra-style consistent, inter-style diverse and high-quality style dataset.
    \item We construct a diverse and balanced prompt gallery containing 170K style prompts and 400K content prompts, yielding up to 68B content–style combinations for training, and we use these prompts to generate the MegaStyle-1.4M dataset.
    \item We propose a style-supervised contrastive learning objective to fine-tune a style encoder, MegaStyle-Encoder, which excels at extracting style-specific representations and enables reliable style similarity measurement.
    \item Experiments show that our MegaStyle-FLUX produces stable, well-generalized stylized results and achieves SOTA performance compared with baseline methods.
\end{itemize}

\section{Related Work}

\subsection{Style Datasets}
Early style datasets are usually collected from the Internet. 
For example, WikiArt \cite{phillips2011wiki} contains 80K real-world artworks by 1,119 artists spanning 27 genres.
JourneyDB \cite{sun2024journeydb} crawls 4.4M high-quality user-generated images from Midjourney, along with 300K short personalized style descriptions.
More recently, Style30K \cite{li2024styletokenizer} first adopts a semi-manual pipeline to construct 30K images spanning 1,120 styles by retrieving images with similar styles.
However, these methods use unreliable style similarity measurement during dataset curation, resulting in style pairs with large intra-style discrepancies that are unsuitable for paired supervision.
To improve intra-style consistency, IMAGStyle \cite{xing2024csgo} and OmniStyle-150K \cite{wang2025omnistyle} utilize SOTA style transfer methods to generate stylized images conditioned on the given reference style images.
Specifically, IMAGStyle trains 15k style and content LoRAs \cite{hu2021lora} and generates 210K stylized images via B-LoRA \cite{frenkel2024implicit}.
OmniStyle-150K builds on the 1,000 styles in Style30K and synthesizes 150K stylized images using StyleID \cite{chung2024style}, StyleShot \cite{11165480}, CSGO \cite{xing2024csgo}, ArtFlow \cite{an2021artflow}, AesPANet \cite{hong2023aespa} and CAST \cite{zhang2022domain}.
However, the inter-style diversity, the quality and the intra-style consistency are heavily limited by the unstable performance of current SOTA style transfer methods.
In this paper, we employ VLMs to construct diverse and balanced 170K styles and 400K contents prompts, and leverage Qwen-Image’s consistent T2I style mapping capability to generate the intra-style consistent, inter-style diverse and high-quality style dataset, MegaStyle-1.4M.

\subsection{Image Style Transfer}
With the development of diffusion models in image generation, numerous style transfer methods have exhibited remarkable performance.
For example, methods \cite{zhou2025attention,jeong2023training, hamazaspyan2023Diffusion, wu2023uncovering, hertz2023style,yang2023zero,chen2023controlstyle,zhangalignedgen} identify style in the feature space of a pre-trained diffusion model and perform editing as training-free style transfer, but with reduced and unstable transfer performance.
Another line of work, tuning-based methods \cite{everaert2023Diffusion, lu2023specialist, gal2022image, zhang2023inversion} fine-tune additional components—such as adapters \cite{sohn2024styledrop,ruiz2023Dreambooth}, text embeddings \cite{zhang2023inversion,gal2022image,voynov2023p+}, or blocks \cite{hu2021lora}—to learn a single style concept from a few style images.
More effectively, recent works \cite{wang2023styleadapter,ahn2024dreamstyler} adapt a pre-trained image encoder (usually CLIP) as a style encoder to extract style features and inject them into a pre-trained diffusion model via cross-attention modules.
These methods are difficult to decouple style from content under the self-supervised training paradigm, often leading to content leakage and inferior style transfer performance.
To address this, some approaches \cite{wang2025omnistyle,xing2024csgo} generate style pairs (i.e., samples that share the same style but differ in content) using SOTA style transfer methods to conduct paired supervision. 
However, the inter-style diversity, the quality, and intra-style consistency of style pairs are constrained by the performance of the style transfer methods used in data curation pipelines, making it difficult to achieve stable and generalizable style transfer performance.
In our work, we use paired supervision to train a FLUX-based style transfer model on MegaStyle-1.4M, enabling stable and generalizable style transfer.

\subsection{Style Similarity Measurement}
Style similarity in image style transfer is often quantified by measuring the distance between the stylized outputs and the provided reference style image.
These distances are typically computed in feature spaces from different models.
Specifically, Gram loss \cite{gatys2016image,huang2017arbitrary} measures the distance between Gram matrices computed from feature maps of a pre-trained CNN model (e.g., VGG \cite{simonyan2014very}).
FID \cite{heusel2017gans} and ArtFID \cite{wright2022artfid} calculate the distribution distance to measure the global style similarity between two style image sets.
Many studies \cite{wang2023styleadapter,qi2024deadiff,ahn2024dreamstyler} utilize CLIP image score to gauge the style similarity in the CLIP's feature space.
However, recent works \cite{sohn2024styledrop, liu2023stylecrafter, 11165480} indicate that these metrics are not ideal for evaluating style similarity, because they rely on feature spaces that are more semantic in nature and are not specialized for capturing style.
To address this, CSD \cite{somepalli2024measuring} fine-tunes the CLIP image encoder on style pairs under style labels from artists, mediums, and movements. 
But with these coarse labels, images in the same style would exhibit large intra-style discrepancies, which can lead to ambiguous style representations and unreliable style evaluation results.
In contrast, we propose a novel style-supervised contrastive learning objective to train MegaStyle-Encoder on MegaStyle-1.4M for more reliable style similarity measurement.

\begin{figure*}[t]
  \includegraphics[width=\textwidth]{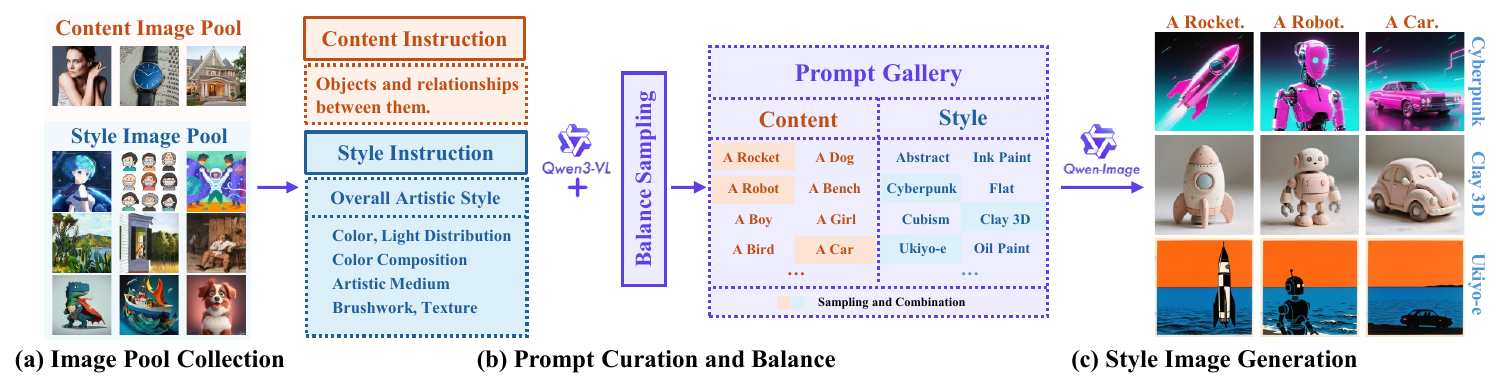}
  \vspace{-8mm}
  \caption{Overview of our data curation pipeline. We first collect style and content images from open-source datasets. Next, we apply carefully designed instructions to generate style and content prompts with Qwen3-VL, together with balance sampling. Finally, we use Qwen-Image to generate style images using content-style prompt combinations. \textbf{Please note that we use simplified content and style prompts for illustrative purposes only}.}
    \vspace{-6mm}
  \label{figs:pipeline}
\end{figure*}

\section{MegaStyle}
In this section, we first introduce the data curation pipeline in Section \ref{sec:3.1}. 
We then present the style-supervised contrastive learning objective and training details of style encoder, MegaStyle-Encoder in Section \ref{sec:3.2}. 
Finally, we introduce MegaStyle-FLUX, our FLUX-based style transfer model, in Section \ref{sec:3.3}.

\subsection{MegaStyle-1.4M}
\label{sec:3.1}
We illustrate our dataset curation pipeline in Figure \ref{figs:pipeline}, which consists of three main stages: Image Pool Collection, Prompt Curation and Balance, and Style Image Generation.

\noindent{\textbf{Image Pool Collection.}} 
We build the content and style image pools from several open-source datasets \cite{phillips2011wiki}. 
Specifically, the style image pool contains 2M images, including about 1M images from a large-scale deduplicated stylized-image corpus, 80K images from WikiArt \cite{phillips2011wiki} covering diverse real-world painting styles, and about 1M additional stylized images filtered from public image collections using style descriptors derived from WikiArt.
For the content image pool, we collect another 2M images from public image collections excluding those used for the style image pool, i.e., the remaining non-stylized images. 
These images span a wide range of visual styles and semantic contents, providing sufficiently diverse style and content priors for subsequent prompt curation.


\begin{figure}
    \centering
    \includegraphics[width=1.0\linewidth]{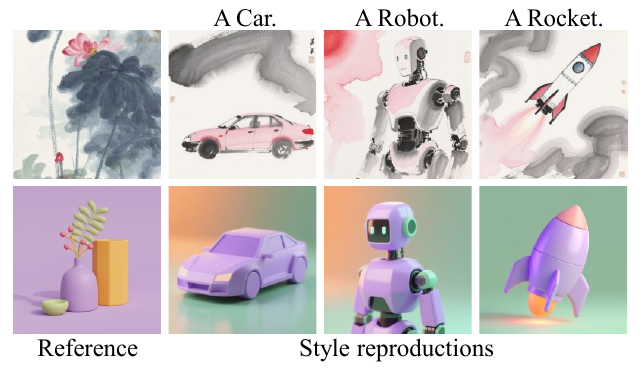}
    \vspace{-6mm}
    \caption{Visualizations of style reproductions. We first use Qwen3-VL to caption a style prompt from the reference style image, and then generate style reproductions on content–style combinations using Qwen-Image.}
    \vspace{-6mm}
    \label{fig:reproduce}
\end{figure}

\noindent{\textbf{Prompt Curation and Balance.}} 
After obtaining the content and style image pools, we generate captions for these images using the powerful VLM Qwen3-VL \cite{Qwen3-VL}, guided by specialized textual instructions for content and style.
We first instruct Qwen3-VL to characterize the style of the input image with an overall artistic style description and several key aspects like color composition and distribution, light distribution, artistic medium, texture, and brushwork, while ignoring the content-related information in the input image.
This formulation of style, together with Qwen3-VL’s strong capabilities, is sufficient to establish an image-to-text style mapping.
As shown in Figure \ref{fig:reproduce}, the style reproductions generated using the style prompts captioned from the reference style images exhibit similar style (ink painting and 3D) with corresponding reference style images.
Please note that these style images should not be regarded as the final style transfer results, as some loss of stylistic detail is inevitable during reproduction.
For the content part, we refer to the instruction prompt used in Qwen-Image, which describes only the objects and their visual relationships, while excluding any style-related descriptions.
This results in a curated prompt gallery of 2M content and style prompts that guarantees a diverse distribution.

We then sample a balanced prompt subset using a two-stage sampling strategy.
We implement the first stage by employing Exact Deduplication, Fuzzy Deduplication and Semantic Deduplication from Nemo-Curator to eliminate exact, near, and semantic duplicates in the prompt gallery, leaving 1M prompts.
\begin{figure}
    \centering
    \includegraphics[width=0.85\linewidth]{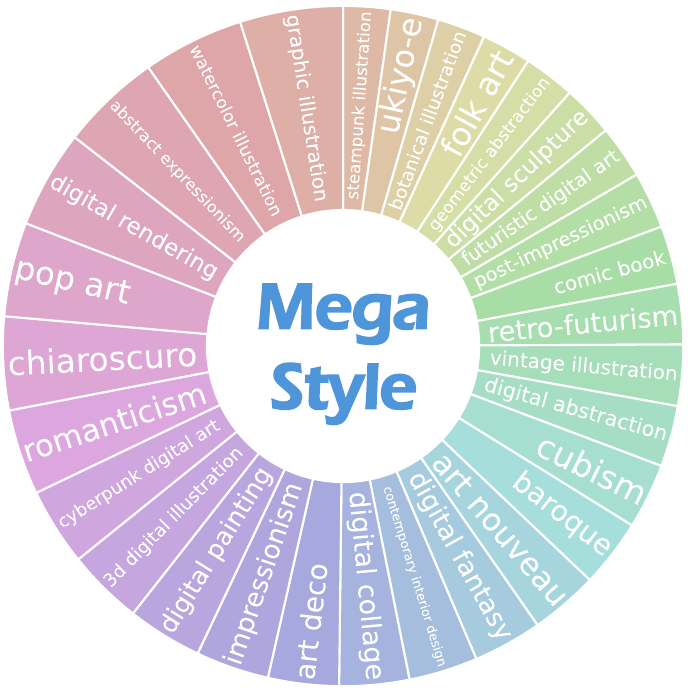}
    \vspace{-2mm}
    \caption{Distribution analysis of overall artistic styles in the style prompts. We present the proportions of the top 30 overall artistic styles.}
    \vspace{-2mm}
    \label{fig:distribution}
\end{figure}
For the second stage, we follow DINOv3 \cite{simeoni2025dinov3}, which applies a balance sampling algorithm based on hierarchical k-means \cite{vo2024automatic} to balance the remaining prompts.
We utilize mpnet \cite{NEURIPS2020_c3a690be} for text embeddings and perform four-level hierarchical clustering with 50K, 10K, 5K, and 1K clusters from the lowest to the highest level. 
This process yields 170K style prompts and 400K content prompts.
We further present a detailed analysis of the overall artistic styles in the style prompts. 
We observe that there are 8K overall artistic style descriptors and we illustrate the proportion of the top 30 styles in Figure \ref{fig:distribution}.
This diverse style distribution is balanced, which benefits our model in learning expressive and generalized style representations.
More details are provided in the supplementary material.

\begin{table}[h]
    \centering
    \setlength{\tabcolsep}{1pt} 
    \setlength{\arrayrulewidth}{0.7pt}
    \caption{Comparison of style datasets. \cmark/\xmark \ indicate whether intra-style consistency is provided and \textemdash\ indicates that the statistic is unavailable.}
    \vspace{-2mm}
    \resizebox{\linewidth}{!}{
    \begin{tabular}{c|cccc}
        \toprule
       Datasets  & \makecell{Intra-style\\Consistency} & \makecell{Overall\\Style} & \makecell{Fine-grained\\Style} & \makecell{Style Image\\Number} \\
       \midrule
       WikiArt  & \xmark & 27 & \textemdash & 80K \\
       JourneyDB  & \xmark & \textemdash & 300K & 4.4M \\
       Style30K  & \xmark & \textemdash & 1K & 30K \\
       \midrule
       IMAGStyle  & \cmark & 14 & 15K & 210K \\
       OmniStyle-150K  & \cmark & \textemdash & 1K & 150K \\
       \textbf{MegaStyle-1.4M}  & \cmark & 8,355 & 170K & 1.4M \\
       \bottomrule
    \end{tabular}
    }
    \vspace{-2mm}
    \label{tab:data}
\end{table}

\begin{figure*}[t]
    \centering
    \includegraphics[width=0.95\linewidth]{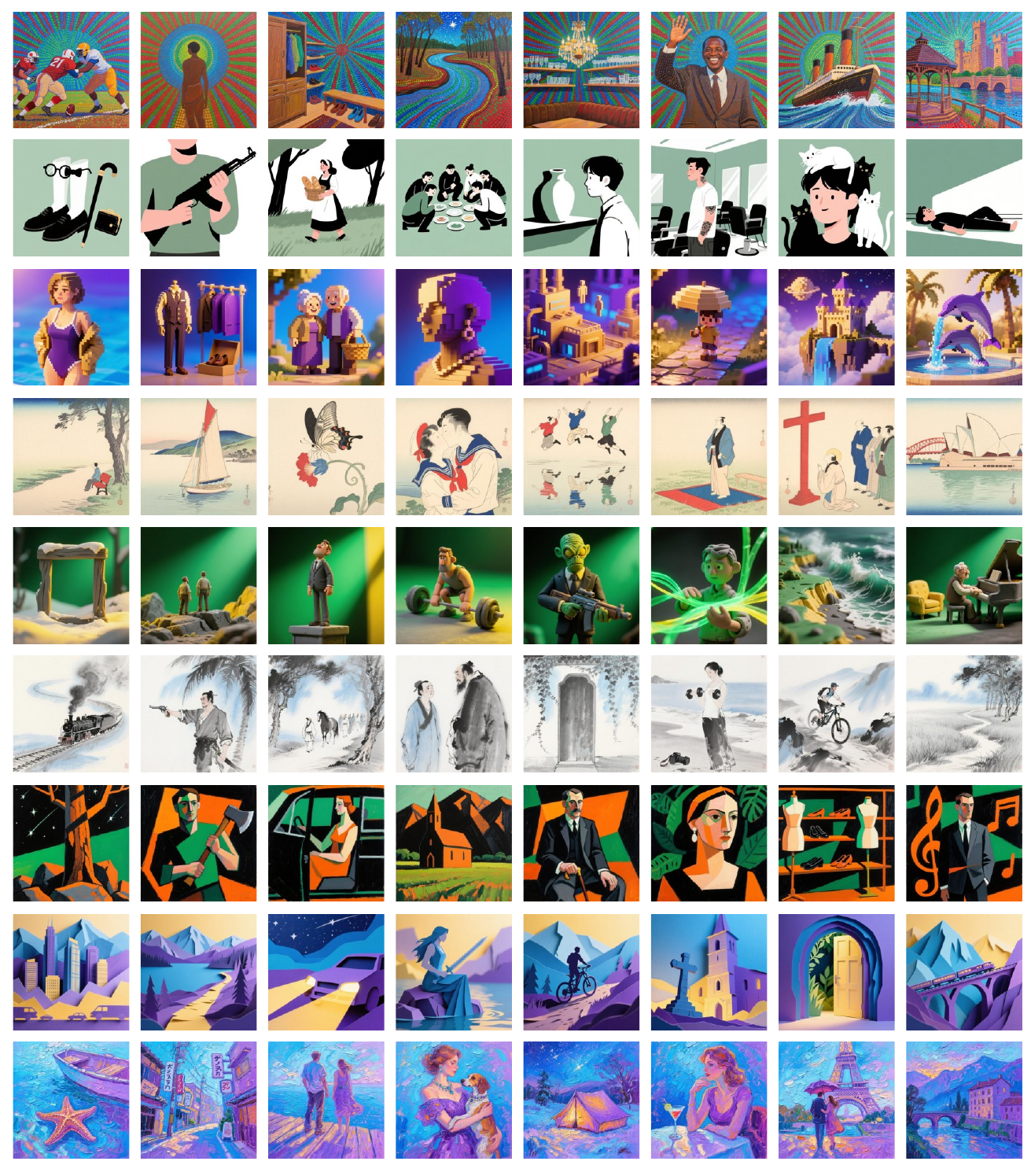}
    \vspace{-3mm}
    \caption{Visualizations of style pairs in MegaStyle-1.4M, where each row shows the same style with different contents.}
    \vspace{-4mm}
    \label{fig:main_dataset}
\end{figure*}

\noindent{\textbf{Style Image Generation.}}
Building on these content and style prompts, we generate style images using Qwen-Image.
Specifically, for each style prompt, we randomly sample $N$ content prompts to form $N$ content–style combinations and synthesize $N$ images that share the same style but contain different content.
We finally generate 1.4M style images, forming the MegaStyle-1.4M for subsequent training.
Table \ref{tab:data} summarizes the comparisons between MegaStyle-1.4M and existing style datasets, including WikiArt \cite{phillips2011wiki}, JourneyDB \cite{sun2024journeydb}, Style30K \cite{li2024styletokenizer}, IMAGStyle \cite{xing2024csgo} and OmniStyle-150K \cite{wang2025omnistyle}.
MegaStyle-1.4M achieves high intra-style consistency while offering a large number of overall artistic styles and diverse fine-grained style categories among the compared datasets.
More importantly, it can be readily scaled to much larger datasets while preserving inter-style diversity, intra-style consistency and high-quality, since each component of MegaStyle’s data curation pipeline is itself scalable, demonstrating strong potential to support broader community research in style transfer and style representation.
Visualizations of style images in MegaStyle-1.4M are presented in Figure \ref{fig:main_dataset} and the supplementary material, the generated images from the same style prompt exhibit strong intra-style consistency.

\subsection{MegaStyle-Encoder}
\label{sec:3.2}
Previous methods \cite{wang2023styleadapter,liu2023stylecrafter,qi2024deadiff} often utilize the image encoder of VLMs to extract style embeddings for style similarity measurement.
However, \cite{11165480} indicates that these image encoders are typically trained with image–text contrastive objectives and the paired texts mainly describe semantic content; and they are more effective at semantic alignment than at modeling image style.
Therefore, leveraging MegaStyle-1.4M, which provides intra-style consistent, inter-style diverse and high-quality style pairs, we propose style-supervised contrastive learning (SSCL) to fine-tune a style encoder (MegaStyle-Encoder) for extracting style-specific representations.

For the image/style-prompt pairs ${(x_k, s_k)}_{k=1}^{MN}$ in MegaStyle-1.4M, where $M$ denotes 170K fine-grained styles, we follow supervised contrastive learning (SCL) \cite{khosla2020supervised} and define the training objective $\mathcal{L}_{\mathrm{scl}}$ as:
{\footnotesize
\begin{equation}
    \mathcal{L}_{\mathrm{scl}}
    = \frac{1}{MN}\sum_{i=1}^{MN}
    \left(
    -\frac{1}{|\mathcal{P}(i)|}\sum_{p\in\mathcal{P}(i)}
    \log
    \frac{\exp\!\left(\mathbf{z}_i^\top \mathbf{z}_p / \tau\right)}
    {\sum_{a\in\mathcal{A}(i)} \exp\!\left(\mathbf{z}_i^\top \mathbf{z}_a / \tau\right)}
    \right),
    \label{eq:scl}
\end{equation}
}
where $z_i=\frac{\mathcal{E_\theta}(x_i)}{||\mathcal{E_\theta}(x_i)||_2}$ represents the $\ell_2$-normalized latent feature of the anchor sample $x_i$ extracted by the image encoder $\mathcal{E_\theta}$; in our implementation, we use the SigLIP image encoder.
$\tau$ is a scalar temperature parameter.
Positive index $p$ is sampled from $\mathcal{P}(i)=\{\,p\in\{1,\dots,MN\}\mid s_p=s_i\,\}\setminus\{\mathrm{self}(i)\}$, and negative index $a$ is sampled from $\mathcal{A}(i)=\{1,\dots,MN\}\setminus\{\mathrm{self}(i)\}$.
Moreover, we introduce an additional SigLIP image–text contrastive loss $\mathcal{L}_{\mathrm{itc}}$ for regularization:
\begin{equation}
\mathcal{L}_{\mathrm{itc}}
= \frac{1}{MN^2}\sum_{i=1}^{MN}\sum_{j=1}^{MN}
\log\!\left(1+\exp\!\left(-y_{ij}\,\mathbf{z}_i^\top \mathbf{t}_j\right)\right),
\label{eq:itc}
\end{equation}
where $\mathbf{t}_j=\frac{\phi(s_j)}{\|\phi(s_j)\|_2}$ is the $\ell_2$-normalized text embedding of the style prompt extracted by the SigLIP text encoder $\phi$.
$y_{ij}=+1$ if $x_i$ is correctly paired with the style prompt of $s_j$, and $y_{ij}=-1$ otherwise.
Finally, we form style-supervised contrastive learning objective $\mathcal{L}_{\mathrm{sscl}}$ as:
\begin{equation}
\mathcal{L}_{\mathrm{sscl}}
= \mathcal{L}_{\mathrm{scl}} + \mathcal{L}_{\mathrm{itc}}.
\label{eq:sscl}
\end{equation}
During training, we adopt a large batch size 8,192 to provide more challenging and diverse negative samples, preventing the model from relying on trivial cues (e.g., color) and encouraging more discriminative style representations.
And only the parameters of the image encoder $\mathcal{E_\theta}$ are updated.

\subsection{MegaStyle-FLUX}
\label{sec:3.3}
\begin{figure}
    \centering
    \includegraphics[width=1.0\linewidth]{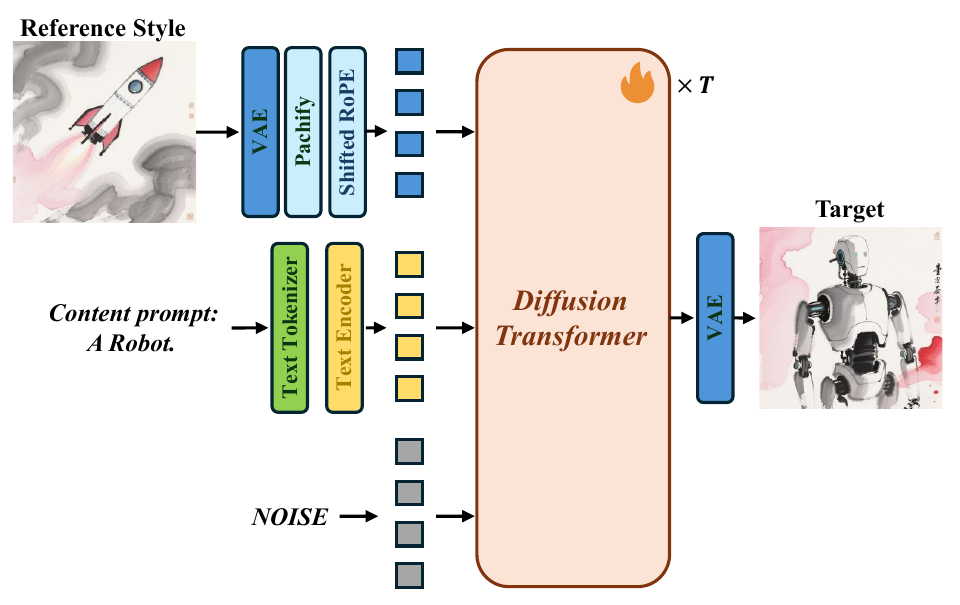}
    \vspace{-6mm}
    \caption{The architecture of MegaStyle-FLUX. }
    \vspace{-6mm}
    \label{fig:flux}
\end{figure}

We build our style transfer model MegaStyle-FLUX on the powerful text-to-image (T2I) model FLUX \cite{flux2024}, the architecture of MegaStyle-FLUX is presented in Figure \ref{fig:flux}.
Specifically, we randomly sample two images sharing the same style from MegaStyle-1.4M, using one as the reference style image and the other as the training target.
The reference style image is encoded and patchified into visual tokens using FLUX’s VAE.
Then we concatenate these reference style tokens
with the noisy image tokens and text tokens and input them into FLUX’s MM-DiT backbone.
We also apply an additional shifted RoPE \cite{zhangalignedgen} to the reference style tokens to prevent positional collision with the target tokens and mitigate cross-image attention bias and content leakage.
During training, we update only the parameters of the diffusion transformer, keep all other components frozen, and use the target image’s content description as the text prompt.
Based on the proposed MegaStyle-1.4M dataset, MegaStyle-FLUX enables generalizable and stable style transfer, faithfully aligning the style of the reference image with the content specified by the text prompt.

\section{Experiments}
\section{Implementation Details}
\noindent{\textbf{Evaluation Metrics.}}
To evaluate the effectiveness of MegaStyle-Encoder in extracting style-specific representations, we follow CSD \cite{somepalli2024measuring} by conducting a style retrieval evaluation and reporting mAP@k and Recall@k, where $k=\{1, 10\}$ denotes the number of top-ranked retrieved images used to compute mAP and Recall. 
Moreover, to evaluate the effectiveness of our style transfer model MegaStyle-FLUX, we follow the style evaluation protocols in previous works \cite{liu2023stylecrafter, 11165480, sohn2024styledrop, wang2023styleadapter} and measure text alignment between the generated image and the text description using the CLIP text score \cite{lin2014microsoft}.
For style similarity measurement, we compute the cosine similarity between the stylized images and the reference style images in the MegaStyle-Encoder feature space.
We also conduct a user study to provide a more comprehensive, human-aligned evaluation of text and style alignment.

\noindent{\textbf{Benchmarks.}}
CSD \cite{somepalli2024measuring} uses WikiArt \cite{phillips2011wiki} as a retrieval benchmark to evaluate style encoder.
As noted above, WikiArt categorizes styles by artist names, which can introduce intra-style discrepancies (see Figure \ref{fig:wikiart}) and therefore make WikiArt unsuitable for evaluating style encoders.
To address this, we sample 2,400 fine-grained styles from the top 800 overall artistic styles not used for training, and pair each with 32 content prompts to construct an intra-style consistent benchmark \textbf{StyleRetrieval} using Qwen-Image.
In StyleRetrieval, we randomly select four images per style as queries and use the remaining 28 images as the gallery.
Moreover, we use the 50 images (real-world artworks) and 20 text prompts from the StyleBench benchmark (as used in StyleShot \cite{11165480}) to evaluate the effectiveness of MegaStyle-1.4M and MegaStyle-FLUX.
\begin{figure}
    \centering
    \includegraphics[width=1.0\linewidth]{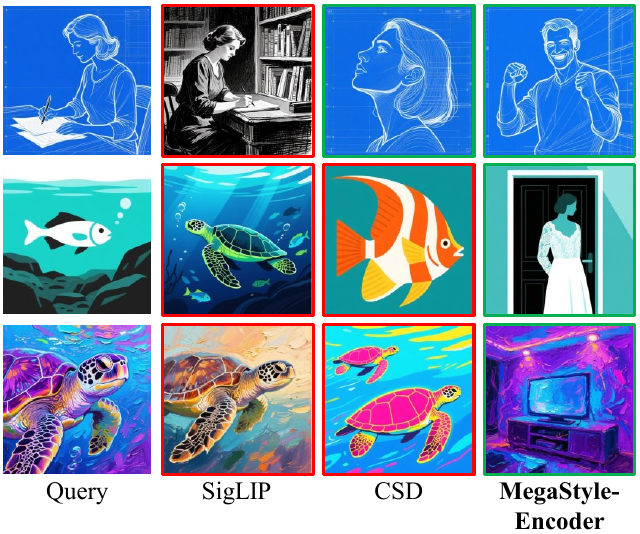}
    \vspace{-6mm}
    \caption{Visual comparison of top-1 matched style retrieval results between MegaStyle-Encoder, SigLIP, and CSD. The red and green borders indicate incorrect and correct matches, respectively.}
    \vspace{-6mm}
    \label{fig:retrieval}
\end{figure}

\begin{table}[h]
    \centering
    \setlength{\tabcolsep}{1pt} 
    \setlength{\arrayrulewidth}{0.7pt}
    \caption{Comparison of MegaStyle-Encoder with other style encoders on the StyleRetrieval benchmark. The best results are highlighted in \textbf{bold}.}
    \resizebox{\linewidth}{!}{
    \begin{tabular}{cc|cc|cc}
        \toprule
         &  & \multicolumn{2}{c}{mAP@k $\uparrow$}  & \multicolumn{2}{|c}{Recall@k $\uparrow$} \\
        \midrule
       Methods  & Backbone & 1 & 10 &1 &10\\
       \midrule
       CLIP   & ViT-L & 9.29 & 6.46 & 9.29& 31.56 \\
       CSD & ViT-L & 45.60 & 37.78 &45.60 & 79.18\\
       MegaStyle-Encoder & ViT-L & \textbf{87.26} & \textbf{85.98} & \textbf{87.26} & \textbf{97.61} \\
       \midrule
       SigLIP  & SoViT & 10.43 & 7.83 &10.43 & 36.32 \\
       MegaStyle-Encoder  & SoViT & \textbf{88.46} & \textbf{86.77} & \textbf{88.46} & \textbf{97.66} \\
       \bottomrule
    \end{tabular}
    }
    \vspace{-2mm}
    \label{tab:retrieval}
\end{table}

\begin{figure*}[t]
    \centering
    \includegraphics[width=1.0\linewidth]{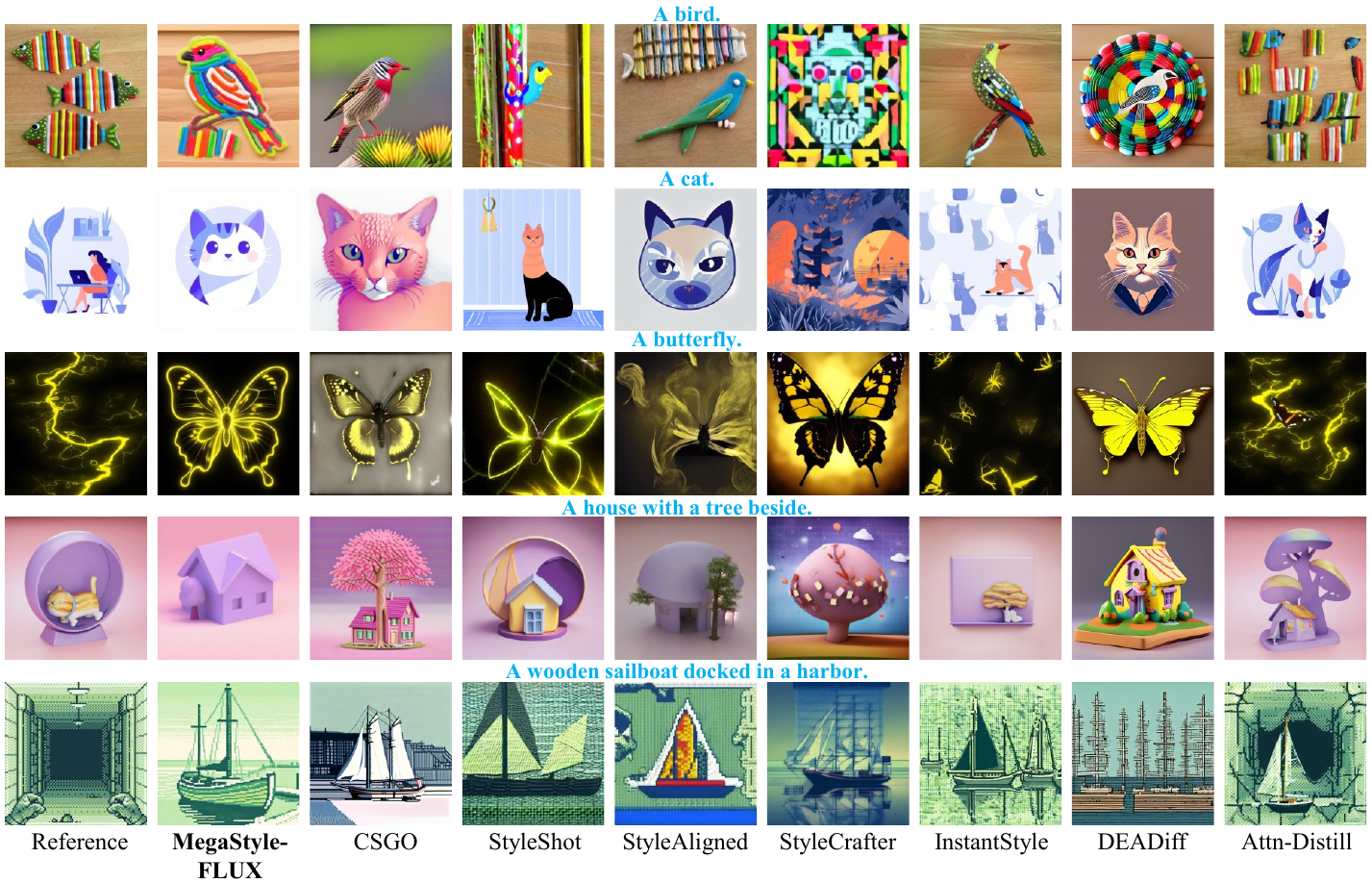}
    \vspace{-6mm}
    \caption{Qualitative comparison between MegaStyle-FLUX and SOTA style transfer methods. MegaStyle-FLUX achieves the superior performance compared to baseline methods.}
    \label{fig:main_result}
\end{figure*}

\begin{table*}[t]
\vspace{-2mm}
\caption{Quantitative comparison of style and text alignment with SOTA style transfer methods. \textit{Style} and \textit{Text} denote the cosine distance in the feature spaces of MegaStyle-Encoder and CLIP, respectively. The prefix \textit{Human} indicates human preference scores. Best result is marked in \textbf{bold}, and the second-best result is highlighted in \underline{underline}.}
\centering
\setlength{\arrayrulewidth}{0.7pt}
\setlength{\tabcolsep}{2.5pt} 
\begin{tabular}{c|cccccccc}
\toprule
{Metrics} & StyleCrafter & DEADiff & Attn-Distill & InstantStyle & CSGO & StyleAligned & StyleShot & MegaStyle-FLUX   \\
\midrule
Style $\uparrow$  & 48.59 & 51.34 & \textbf{85.59} & 71.41 & 55.02 & 59.80 & 63.42 & \underline{76.16}  \\
Text $\uparrow$ & 21.39 & \underline{23.13} & 20.29 & 20.77 & 23.05 & 21.31  & 21.79 & \textbf{23.20} \\
\midrule
Human Style $\uparrow$ & 3.41 & 3.05 & 13.97 &\underline{18.19} & 7.34& 7.46 & 15.21 & \textbf{31.37}\\
Human Text $\uparrow$ & 8.87 & 11.13 & 6.31 & 10.98 & \underline{16.18}& 4.12& 13.69 & \textbf{28.72}\\
\bottomrule
\end{tabular}
\vspace{-2mm}
\label{tab:main_result}
\end{table*}

\subsection{Style Similarity Measurement}
We compare our style encoder MegaStyle-Encoder with the recent style encoder CSD \cite{somepalli2024measuring}, as well as with other VLMs such as CLIP \cite{radford2021learning} and SigLIP \cite{zhai2023sigmoid} on StyleRetrieval.
For a fair comparison, we additionally implement a ViT-L–based MegaStyle-Encoder to match the backbone used by CLIP and CSD.
As shown by the quantitative results in Table \ref{tab:retrieval}, our MegaStyle-Encoder achieves substantially higher mAP and Recall scores than all other methods across all backbones, with a large margin.
We also visualize the top-1 matched image for each query style image of the CSD, SigLIP and MegaStyle-Encoder.
As shown in Figure \ref{fig:retrieval}, for a given query style image, the most similar image retrieved by SigLIP is often biased toward semantic content rather than style.
CSD performs better than SigLIP, but it still relies on content cues for style matching.
We attribute this to the coarse style labels in its training dataset, where style pairs within a style may share similar content and exhibit intra-style discrepancy.
In contrast, our MegaStyle-Encoder accurately retrieves the correct style for each query even when no content is shared, demonstrating its ability to extract expressive, style-specific representations and provide reliable style similarity measurement.

\begin{figure}
\centering
    \includegraphics[width=1.0\linewidth]{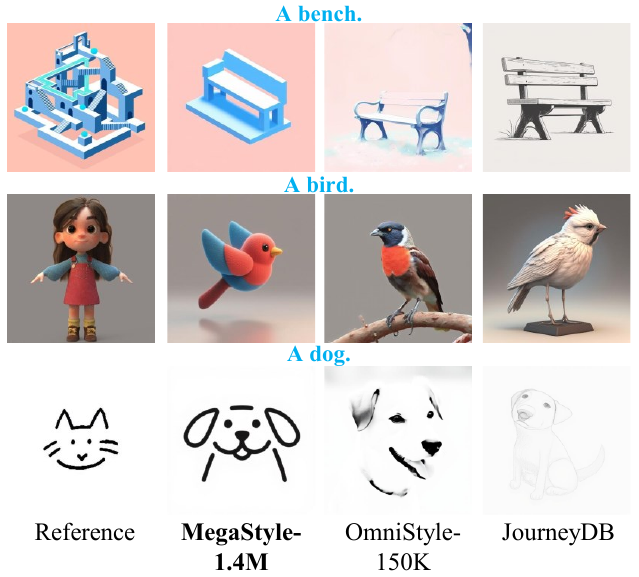}
    \vspace{-6mm}
    \caption{Visual results of MegaStyle-FLUX trained on different style datasets.}
    \vspace{-6mm}
    \label{fig:dataset}
\end{figure}

\subsection{Style Transfer}
We compare MegaStyle-FLUX with the SOTA style transfer methods, including DEADiff \cite{qi2024deadiff}, StyleShot \cite{11165480}, Attention-Distillation (Attn-Distill) \cite{zhou2025attention}, CSGO \cite{xing2024csgo}, StyleCrafter \cite{liu2023stylecrafter}, InstantStyle \cite{wang2024instantstyle} and StyleAligned \cite{hertz2023style}.
We first present visualizations in Figure \ref{fig:main_result}.
Since they were trained on a dataset with limited styles, CSGO, DEADiff, and StyleCrafter exhibit the poor performance on these styles, often transferring only the basic colors from the reference style images.
StyleShot and StyleAligned perform better but content leakage occurs (e.g., the disc in row 4). 
We also observe that InstantStyle and Attention-Distillation respond poorly to the text prompt and tend to copy the reference image (e.g., the clay strip in row 1 and the leaves in row 2).
In contrast, MegaStyle-FLUX generates stylized images that align with the content specified by the text prompt and the style of the reference image.
The quantitative results in Table \ref{tab:main_result} also support these observations.
StyleCrafter, DEADiff, and CSGO have the lowest style alignment scores.
StyleShot and StyleAligned attain relatively high style alignment scores but lower text-alignment scores, due to content leakage.
By largely copying the reference image, Attention-Distillation and InstantStyle achieve very high style alignment scores yet the lowest text alignment scores.
MegaStyle-FLUX achieves the highest text alignment score, the second-best style alignment score, and the highest human preference scores, demonstrating its stable and generalizable performance.
More visual results are shown in the supplementary material.

\subsection{Ablation Studies}

\noindent\textbf{Style Datasets.} To evaluate the effectiveness of our proposed style dataset MegaStyle-1.4M, we compare it with other style datasets like OmniStyle-150K \cite{wang2025omnistyle} and JourneyDB \cite{sun2024journeydb} by training MegaStyle-FLUX on each dataset.
As shown in Figure \ref{fig:dataset}, the model trained on OmniStyle-150K only transfers the basic color of the reference style due to the limited styles in training dataset.
Moreover, the model trained on JourneyDB even fails to capture the colors of the reference style image because the training pairs exhibit inconsistent styles.
With MegaStyle-1.4M, the model performs well across various styles, highlighting the importance of maintaining intra-style consistency in constructing large-scale style datasets.
We also observe that the model trained on MegaStyle-1.4M achieves the best scores in Table \ref{tab:dataset}, further demonstrating its effectiveness.

\begin{figure}[t]
\centering
    \includegraphics[width=1.0\linewidth]{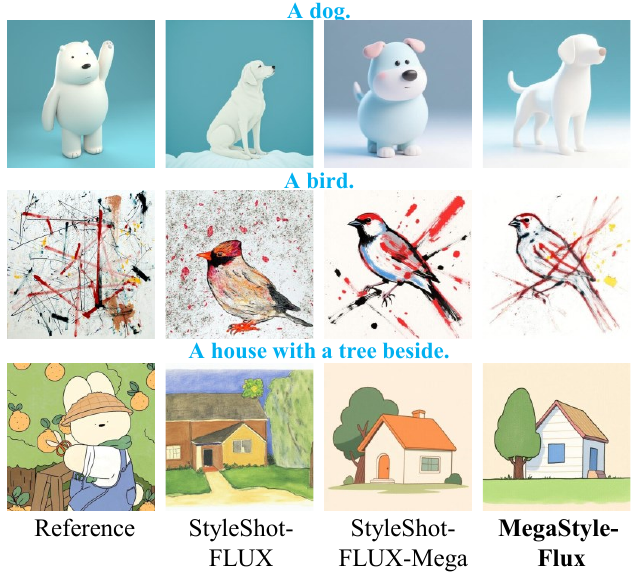}
    \vspace{-6mm}
    \caption{Visual results of MegaStyle-FLUX and fine-tuned StyleShot.}
    \vspace{-2mm}
    \label{fig:styleshot}
\end{figure}

\begin{table}[h]
\caption{Quantitative comparison of style datasets. Best is marked in \textbf{bold}.}
\centering
\setlength{\arrayrulewidth}{0.7pt}
\setlength{\tabcolsep}{3pt} 
\resizebox{\linewidth}{!}{
\begin{tabular}{c|ccc}
\toprule
{Metrics} & JourneyDB & OmniStyle-150K & MegaStyle-1.4M    \\
\midrule
Style $\uparrow$  & 34.56 & 51.49 &  \textbf{76.16}  \\
Text $\uparrow$ &  21.12& 23.02 & \textbf{23.20} \\
\bottomrule
\end{tabular}
}
\vspace{-2mm}
\label{tab:dataset}
\end{table}

\noindent\textbf{Style Encoders.} In our implementation, we use StyleRetrieval as a benchmark to evaluate style encoders.
Although the style pairs in StyleRetrieval exhibit high intra-style consistency, they are generated by the same model (Qwen-Image) used to train MegaStyle-Encoder, which may introduce source-model bias into the evaluation.
To further evaluate MegaStyle-Encoder beyond Qwen-Image's distribution, we additionally compare it with commonly used style encoders, including CLIP and CSD, on StyleBench (275 real-world artworks in 40 styles, following StyleShot), FLUX-Retrieval (76,800 images generated by FLUX across 2,400 styles using the prompts from StyleRetrieval), and OmniStyle150K (30,400 images in 950 styles, following OmniStyle), where one image per style is used as the query in StyleBench, and four images per style are used as queries in FLUX-Retrieval and OmniStyle150K. 
Quantitative results in Tables \ref{tab:encoder} show that, although the style pairs in these benchmarks exhibit lower intra-style consistency than those in StyleRetrieval (as evidenced in Figure \ref{fig:intro}), MegaStyle-Encoder still outperforms all other style encoders across all metrics and benchmarks. 
These results further confirm its robustness and generalization to a broader range of artistic styles, including real-world artworks and synthetic images.

\begin{table*}[t]
    \centering
    \vspace{-2mm}
    \setlength{\tabcolsep}{5pt} 
    \setlength{\arrayrulewidth}{0.7pt}
    \caption{Comparison of MegaStyle-Encoder with other style encoders on the StyleBench, FLUX-Retrieval and OmniStyle-150K. The best results are highlighted in \textbf{bold}.}
    \begin{tabular}{c|cc|cc|cc|cc|cc|cc}
        \toprule
        &  \multicolumn{4}{c|}{StyleBench} & \multicolumn{4}{c|}{FLUX-Retrieval} & \multicolumn{4}{c}{OmniStyle-150K}  \\
        \midrule
         &  \multicolumn{2}{c}{mAP@k $\uparrow$}  & \multicolumn{2}{|c|}{Recall@k $\uparrow$} &  \multicolumn{2}{c}{mAP@k $\uparrow$}  & \multicolumn{2}{|c|}{Recall@k $\uparrow$} &  \multicolumn{2}{c}{mAP@k $\uparrow$}  & \multicolumn{2}{|c}{Recall@k $\uparrow$} \\
        \midrule
       Methods  & 1 & 10 &1 &10 & 1 & 10 &1 &10 & 1 & 10 &1 &10\\
       \midrule
       CLIP   & 40.00 & 30.85 & 40.00& 82.50 & 2.42 & 1.55 & 2.42& 9.68 & 1.68 & 1.35 & 1.68& 10.39\\
       CSD &  70.00 & 51.65 &70.00 & 97.50 &  14.16 & 9.91 &14.16 & 40.08 &  60.86 & 48.24 &60.86 & 89.71\\
       MegaStyle-Encoder & \textbf{85.00} & \textbf{54.15} & \textbf{85.00} & \textbf{100.00} & \textbf{22.70} & \textbf{18.38} & \textbf{22.70} & \textbf{51.87} & \textbf{78.89} & \textbf{60.18} & \textbf{78.89} & \textbf{94.07} \\
       \bottomrule
    \end{tabular}
    \vspace{-2mm}
    \label{tab:encoder}
\end{table*}

\noindent\textbf{Style Transfer Models.}
To ensure a fairer comparison between the baseline methods and MegaStyle-FLUX, we train StyleShot\cite{11165480}—the only baseline with available training script—on FLUX with two datasets: its original dataset StyleGallery (StyleShot-FLUX) and MegaStyle1.4M (StyleShot-FLUX-Mega) to match the base setting of MegaStyle-FLUX. 
As shown in Figure \ref{fig:styleshot}, StyleShot-FLUX transfers only basic stylistic attributes from the reference image, such as color. 
When trained on MegaStyle1.4M, StyleShot-FLUX-Mega effectively captures higher-level styles, such as 3D, flat, and ink.
The quantitative results in Table \ref{tab:styleshot} further support this visual evidence, showing that StyleShot-FLUX-Mega outperforms StyleShot-FLUX across all metrics and further demonstrating the effectiveness of MegaStyle-1.4M.
However, StyleShot encodes style reference images through an extra image encoder (SigLIP), which maps them into a high-level feature space and may lose fine-grained style details, leading to worse performance than MegaStyle-FLUX.

\begin{table}[h]
\vspace{-2mm}
\caption{Quantitative comparison between MegaStyle-FLUX and fine-tuned StyleShot. Best is marked in \textbf{bold}.}
\centering
\setlength{\arrayrulewidth}{0.7pt}
\setlength{\tabcolsep}{1pt} 
\resizebox{\linewidth}{!}{
\begin{tabular}{c|ccc}
\toprule
{Metrics} & StyleShot-FLUX & StyleShot-FLUX-Mega & MegaStyle-FLUX    \\
\midrule
Style $\uparrow$  & 57.06 & 67.73 &  \textbf{76.16}  \\
Text $\uparrow$ &  21.86 & \textbf{23.27} & 23.20 \\
\bottomrule
\end{tabular}
}
\vspace{-2mm}
\label{tab:styleshot}
\end{table}
\section{Conclusion}
In this paper, we propose a scalable data curation pipeline MegaStyle that constructs an intra-style consistent, inter-style diverse and high-quality style dataset.
Leveraging the consistent text-to-image style mapping capability of modern large generative models—which can generate images in the same style from a given style description—we curate a diverse and balanced prompt gallery and generate a large-scale style dataset, MegaStyle-1.4M.
With MegaStyle-1.4M, we propose style-supervised contrastive learning to fine-tune MegaStyle-Encoder for reliable style similarity measurement and we train MegaStyle-FLUX for generalizable and stable style transfer.
Extensive experiments demonstrate the effectiveness of our proposed data curation pipeline, dataset and models, offering valuable insights and contributions to the style transfer community.

\noindent\textbf{Future Work.} In captioning style prompts, we observe that VLMs may produce vague words when describing style elements such as texture, brushwork, and medium. 
This likely occurs because our instruction prompt does not specify which visual aspects the VLM should rely on when identifying these elements.
In future work, we will further refine the instruction prompt to better cover a broader style space and scale our style dataset to the 10-million level.

{
    \small
    \bibliographystyle{ieeenat_fullname}
    \bibliography{main}
}

\clearpage
\setcounter{page}{1}
\maketitlesupplementary

\section{Implementation Details}
In the data curation pipeline, we use the powerful VLM Qwen3-VL-30B-A3B-Instruct\footnote{\url{https://huggingface.co/Qwen/Qwen3-VL-30B-A3B-Instruct}} to generate content and style prompts from the collected images, following carefully designed instruction templates, with $N=8$.
In balance sampling, we use all-mpnet-base-v2\footnote{\url{https://github.com/replicate/all-mpnet-base-v2}} for text embedding.
During fine-tuning of the MegaStyle-Encoder, we use siglip-so400m-patch14-384\footnote{\url{https://huggingface.co/google/siglip-so400m-patch14-384}} as the base model and fine-tune it for 30 epochs on MegaStyle-1.4M with a batch size of 8,192, a learning rate of 5e-4, a weight decay of 0.01, and $\tau=0.07$. 
We train our style transfer model, MegaStyle-FLUX, on FLUX.1-dev\footnote{\url{https://huggingface.co/black-forest-labs/FLUX.1-dev}} for 30,000 steps, using a batch size of 8, a learning rate of 1e-4, and a 512×512 resolution, with a LoRA rank of 128.
We use FlowMatchScheduler with 40 inference steps and cfg\_scale = 4.0 during Qwen-Image generation. In balance sampling, we first encode all prompts using mpnet embeddings, and then perform a bottom-up four-level hierarchical k-means with \(k=\{50\text{K},10\text{K},5\text{K},1\text{K}\}\), where the lowest-level clusters the raw embeddings and each higher level clusters the centroids from the previous level. Next, we adopt top-down hierarchical sampling to form the balanced set. For a target budget \(M\), we start from the top level of the hierarchy and use:
$$\arg\min_n \left| M-\sum_j \min(n,s_j)\right|$$
to determine a shared cap \(n\), where \(s_j\) denotes the size of the \(j\)-th cluster, so that \(\min(n,s_j)\) samples are allocated to each cluster at the next lower level. We recursively apply this process until reaching the lowest-level clusters, where the final prompts are sampled. 

\noindent{\textbf{Human Preference.}}
We elaborate on the human preference study reported in Section 4.
We construct 20 evaluation tasks for style transfer to enable controlled comparisons. 
In each task, assessors are shown a reference style image, a text prompt and the corresponding stylizations. 
For every task, we supply clear guidelines and collect judgments from more than 30 volunteers. 
The complete experimental protocol and the instructions are described below.
\begin{promptbox}[User Study]

In our user study, we conducted evaluations on 20 tasks. Each task provided a reference style image along with outputs generated by style transfer methods. Participants were instructed to rank the results based on how closely they aligned with the reference style and text prompt according to the following criteria:
\begin{itemize}
    \item Style Consistency: The style of the generated image aligns with that of the reference style image;

    \item Text Consistency: The depicted content of generated image correspond with the textual description;
\end{itemize}
The questions are as follows:
\begin{itemize}
    \item Please rank the generated images—Image A through Image H—according to how well each matches the style of the reference image.

    \item Please rank the generated images—Image A through Image H—according to how well each matches the description by the text prompt.
\end{itemize}

\end{promptbox}

We assign weighted scores based on the resulting rankings as final scores.

\noindent{\textbf{Instruction Templates.}}
We provide the instruction templates of content and style prompt.
For captioning style prompt, we use:
\begin{promptbox}[Style Caption]

Image Annotator

You are a professional image annotator. Please characterize the style of the input image in 32 words based on the following instructions:

1. Start with an overall artistic style description.

2. Identify and specify only the following style features:

    - color composition and distribution.
    
    - light distribution.
    
    - artistic medium.
    
    - texture from surface roughness, layering, density and reflectivity.
    
    - brushwork from stroke width/length, direction, shape and edge hardness.

    In describing each style feature, do not mention any recognizable subjects, objects, environmental context or natural-scene terms.

3. Avoid starting captions with instructional phrases like "The image", "A figure" etc.

\#\# Output Format

"In the style of \{artistic style\}, \{main color\} with \{other colors\} in \{color distribution\}, \{light distribution\} light, \{artistic medium\}, \{texture\}, \{brushwork\}."

\end{promptbox}

For content prompt, we use:

\begin{promptbox}[Content Caption]

Image Annotator

You are a professional image annotator. Please create the caption for the input image in 64 words based on the following instructions:

1. Describe only the objects and the visual relationships between them using natural text without structured formats or rich text.

2. Maintain authenticity and accuracy; avoid generalizations.

3. Exclude any style-related descriptions, such as color, lighting, texture, brushwork, material characteristics, artistic medium, mood, and artistic style.

4. Avoid starting captions with instructional phrases like "The image", "A figure" etc.

5. Do not include any color descriptions under any circumstances.

\#\# Sample Output Format

"..."

\end{promptbox}
    
\noindent{\textbf{Proportion Values.}}
We also report the proportion of the top 30 overall artist styles in Figure \ref{fig:distribution} as graphic illustration (1.18\%), watercolor illustration (1.16\%), abstract expressionism (1.15\%), digital rendering (1.12\%), pop art (1.08\%), chiaroscuro (1.07\%), Romanticism (0.98\%), cyberpunk digital art (0.89\%), 3D digital illustration (0.87\%), digital painting (0.86\%), impressionism (0.84\%), Art Deco (0.81\%), digital collage (0.80\%), digital fantasy (0.79\%), contemporary interior design (0.79\%), Baroque (0.78\%), Art Nouveau (0.78\%), Cubism (0.75\%), vintage illustration (0.70\%), digital abstraction (0.70\%), retro-futurism (0.69\%), comic book (0.67\%), Post-Impressionism (0.65\%), futuristic digital art (0.61\%), geometric abstraction (0.59\%), digital sculpture (0.59\%), folk art (0.57\%), ukiyo-e (0.55\%), botanical illustration (0.55\%), steampunk illustration (0.54\%).

\begin{figure}[t]
    \centering
    \includegraphics[width=1.0\linewidth]{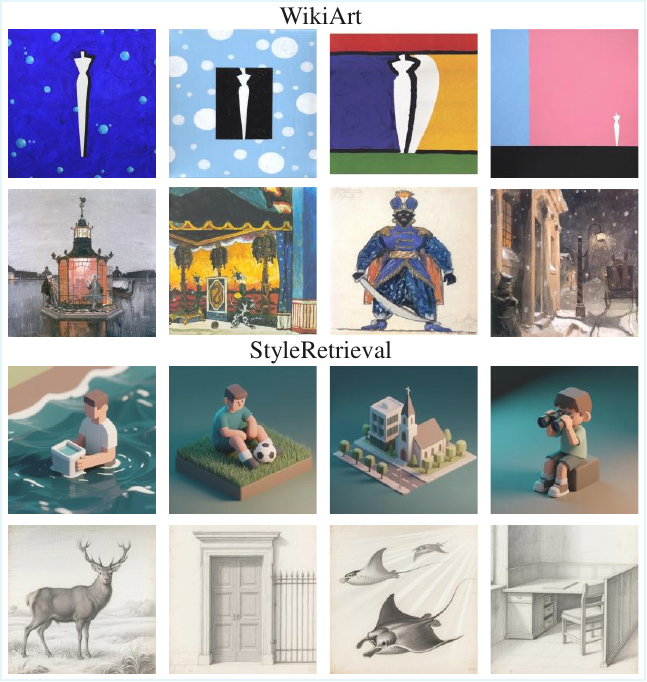}
    \vspace{-6mm}
    \caption{Visualizations of retrieval benchmark WikiArt and our StyleRetrieval, where each row shows the same style during retrieval. }
    \vspace{-6mm}
    \label{fig:wikiart}
\end{figure}
\section{Experiments}
\subsection{Retrieval Benchmark}
In this subsection, we present the visualizations of style retrieval benchmark WikiArt (used in previous methods) and our StyleRetrieval.
As shown in Figure \ref{fig:wikiart}, images in WikiArt exhibit substantial intra-style discrepancies (especially in color, texture and brushwork) because WikiArt categorizes styles by artist names. 
In addition, the image contents are often highly similar (row 1). 
These severely hinder a proper evaluation of the style encoder’s representations and its style retrieval capability.
In contrast, we leverage Qwen-Image’s consistent text-to-image mapping capability to generate images for StyleRetrieval that share the same style but depict different content, making the dataset well-suited for evaluating style encoders.

\subsection{Comparison with Qwen-Image-Edit}
\begin{figure}[t]
    \centering
    \includegraphics[width=0.75\linewidth]{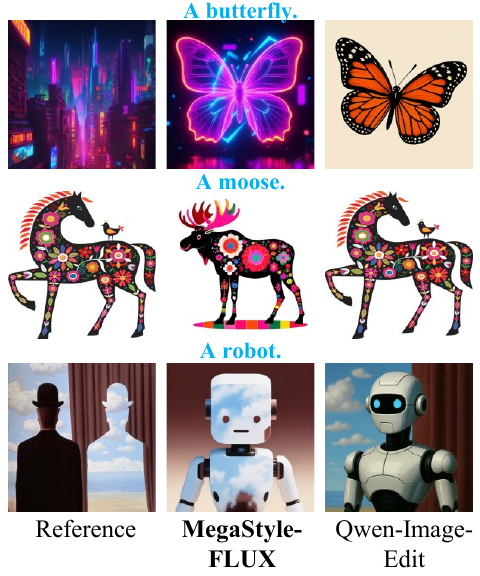}
    \vspace{-2mm}
    \caption{Visual results between MegaStyle-FLUX and Qwen-Image-Edit. }
    \vspace{-2mm}
    \label{fig:qwen}
\end{figure}

We compare MegaStyle-FLUX with Qwen-Image-Edit in Table \ref{tab:qwen} and Figure \ref{fig:qwen}. MegaStyle-FLUX significantly outperforms Qwen-Image-Edit on style transfer. This is likely because Qwen-Image-Edit is primarily trained on editing image pairs, whereas MegaStyle-FLUX is trained on large-scale, high-quality style image pairs, demonstrating the necessity of our proposed MegaStyle-1.4M dataset for training a style transfer model.

\begin{table}[h]
\caption{Quantitative comparison between MegaStyle-FLUX and Qwen-Image-Edit. Best is marked in \textbf{bold}.}
\centering
\setlength{\arrayrulewidth}{0.7pt}
\setlength{\tabcolsep}{7pt} 
\begin{tabular}{c|cc}
\toprule
{Metrics} & Qwen-Image-Edit & MegaStyle-FLUX    \\
\midrule
Style $\uparrow$  & 43.03 &  \textbf{76.16}  \\
Text $\uparrow$ &  \textbf{24.24} & 23.20 \\
\bottomrule
\end{tabular}
\vspace{-2mm}
\label{tab:qwen}
\end{table}

\subsection{More Visualizations}
\begin{figure*}[t]
    \centering
    \includegraphics[width=1.0\linewidth]{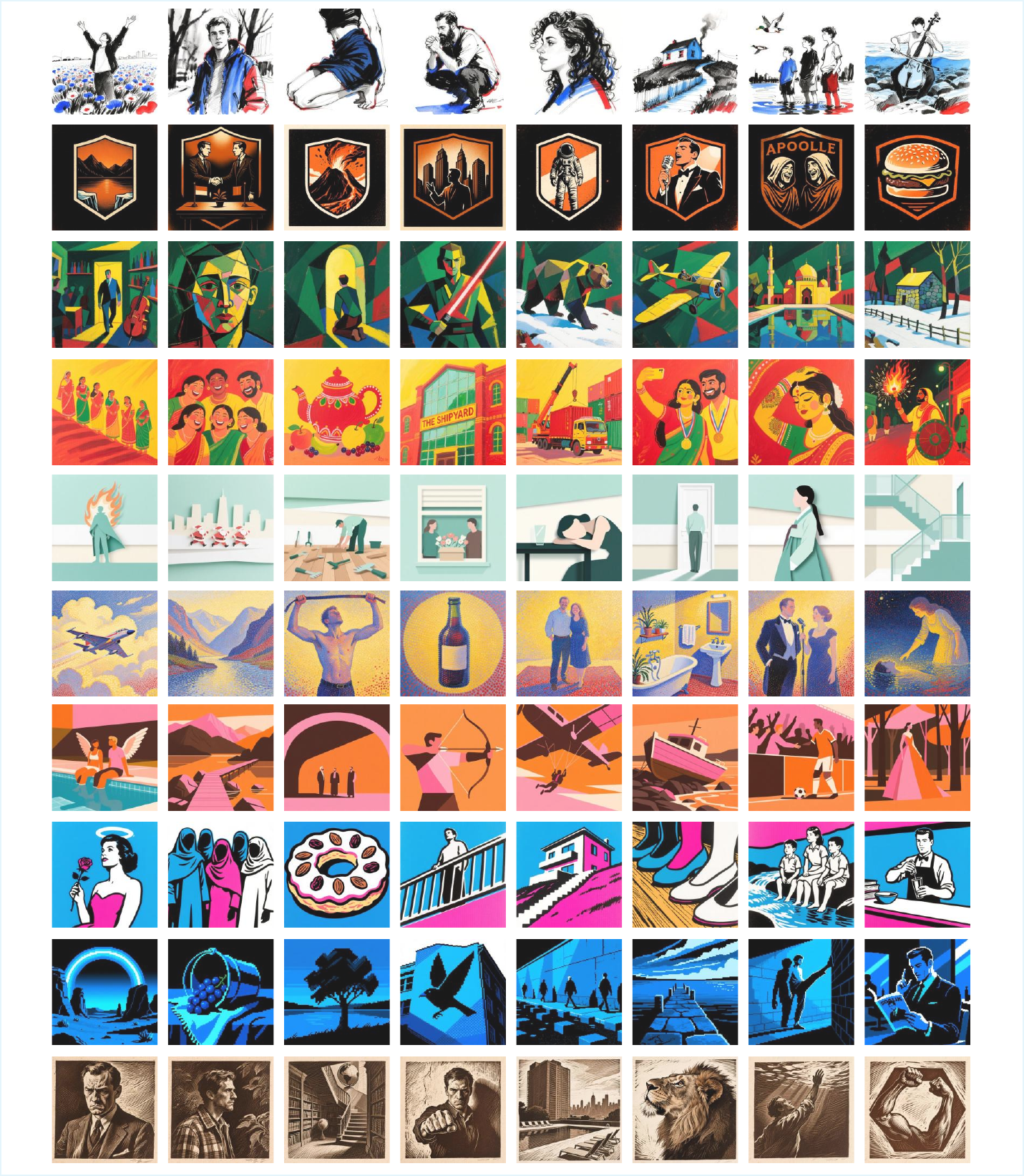}

    \caption{Additional visualizations of style pairs in MegaStyle-1.4M, where each row shows the same style with different contents.}
    \vspace{-4mm}
    \label{fig:main_dataset1}
\end{figure*}

\begin{figure*}[t]
    \centering
    \includegraphics[width=1.0\linewidth]{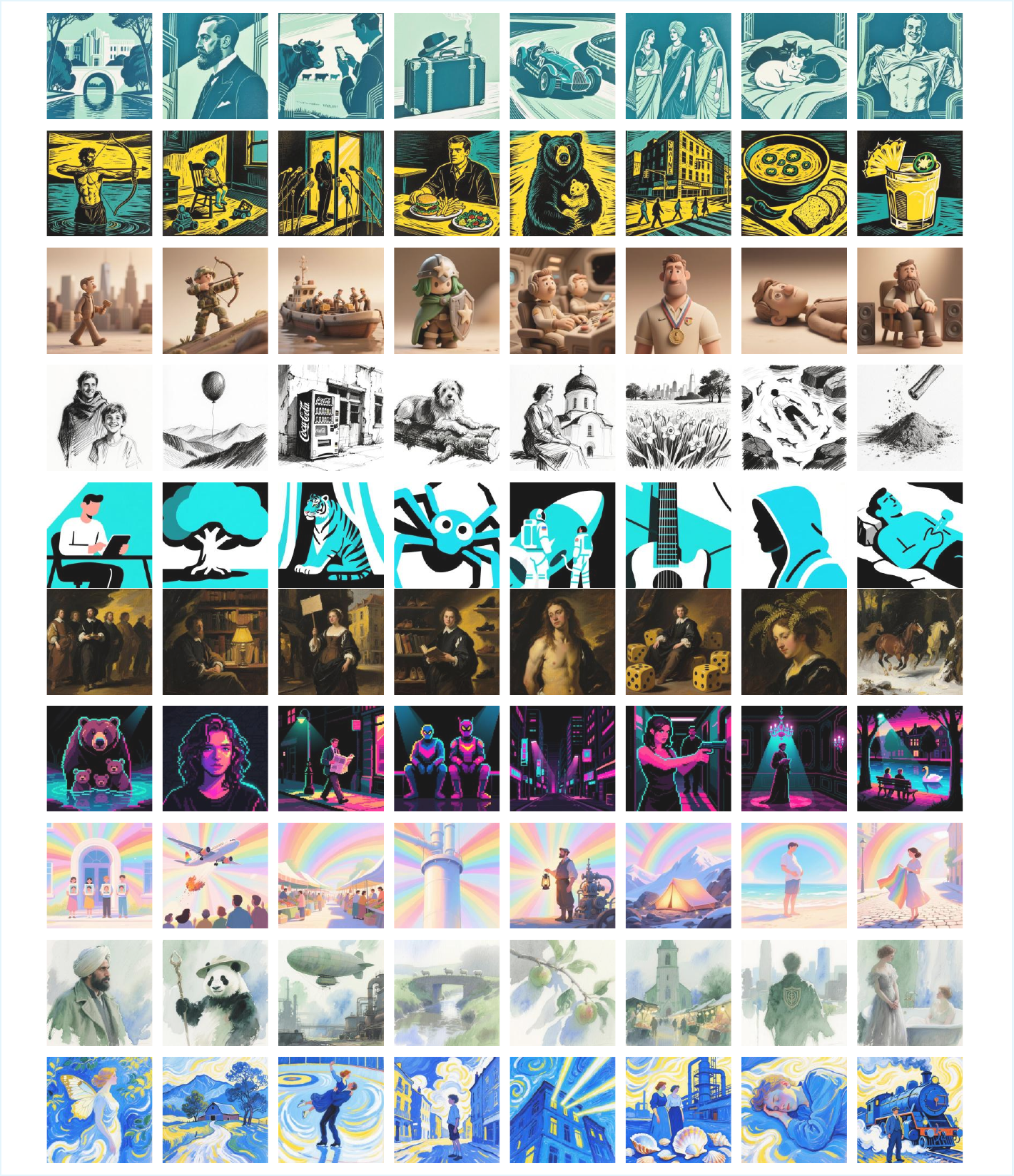}

    \caption{Additional visualizations of style pairs in MegaStyle-1.4M, where each row shows the same style with different contents.}
    \vspace{-4mm}
    \label{fig:main_dataset2}
\end{figure*}

\begin{figure*}[t]
    \centering
    \includegraphics[width=1.0\linewidth]{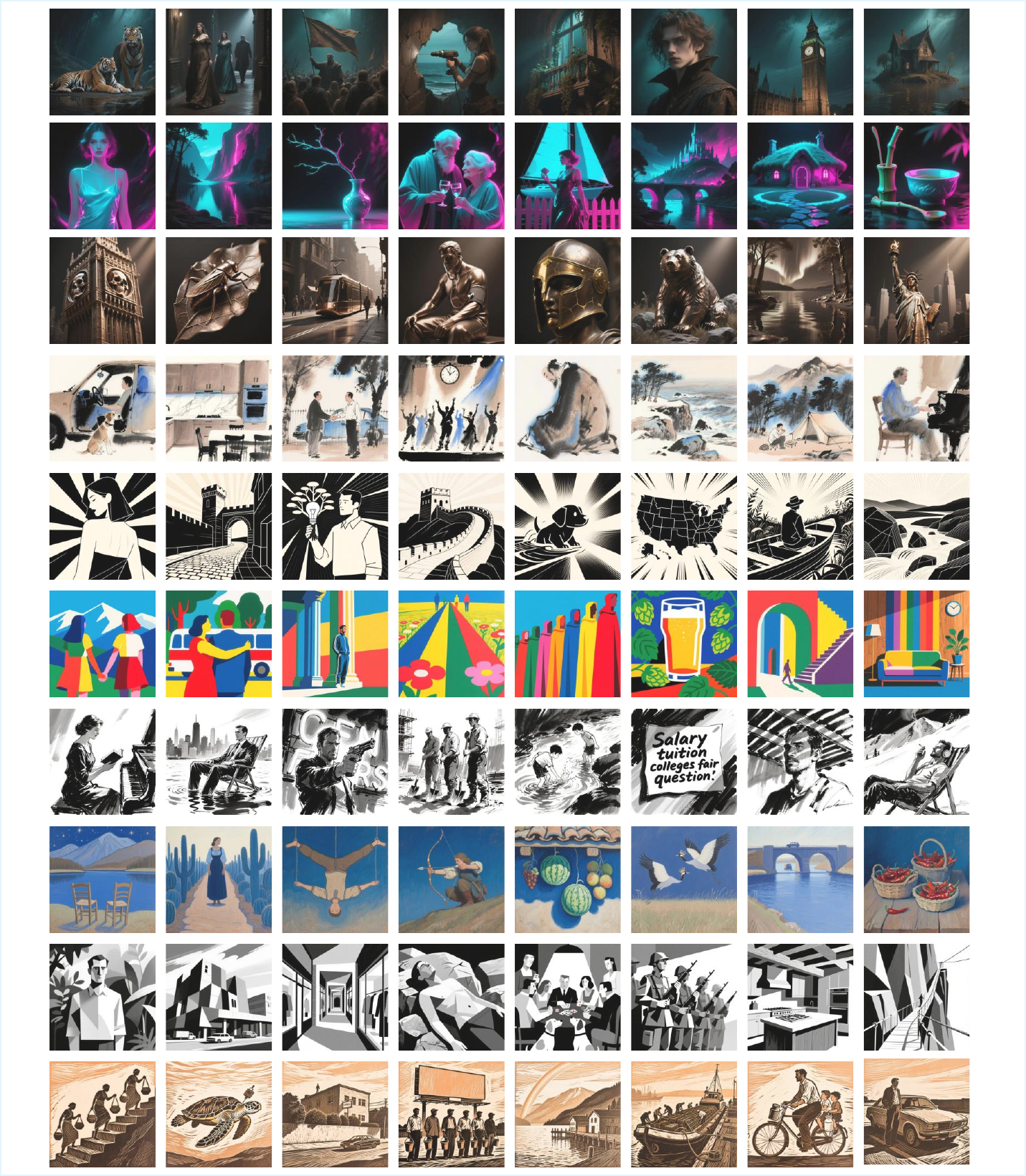}

    \caption{Additional visualizations of style pairs in MegaStyle-1.4M, where each row shows the same style with different contents.}
    \vspace{-4mm}
    \label{fig:main_dataset3}
\end{figure*}

\begin{figure*}[t]
    \centering
    \includegraphics[width=1.0\linewidth]{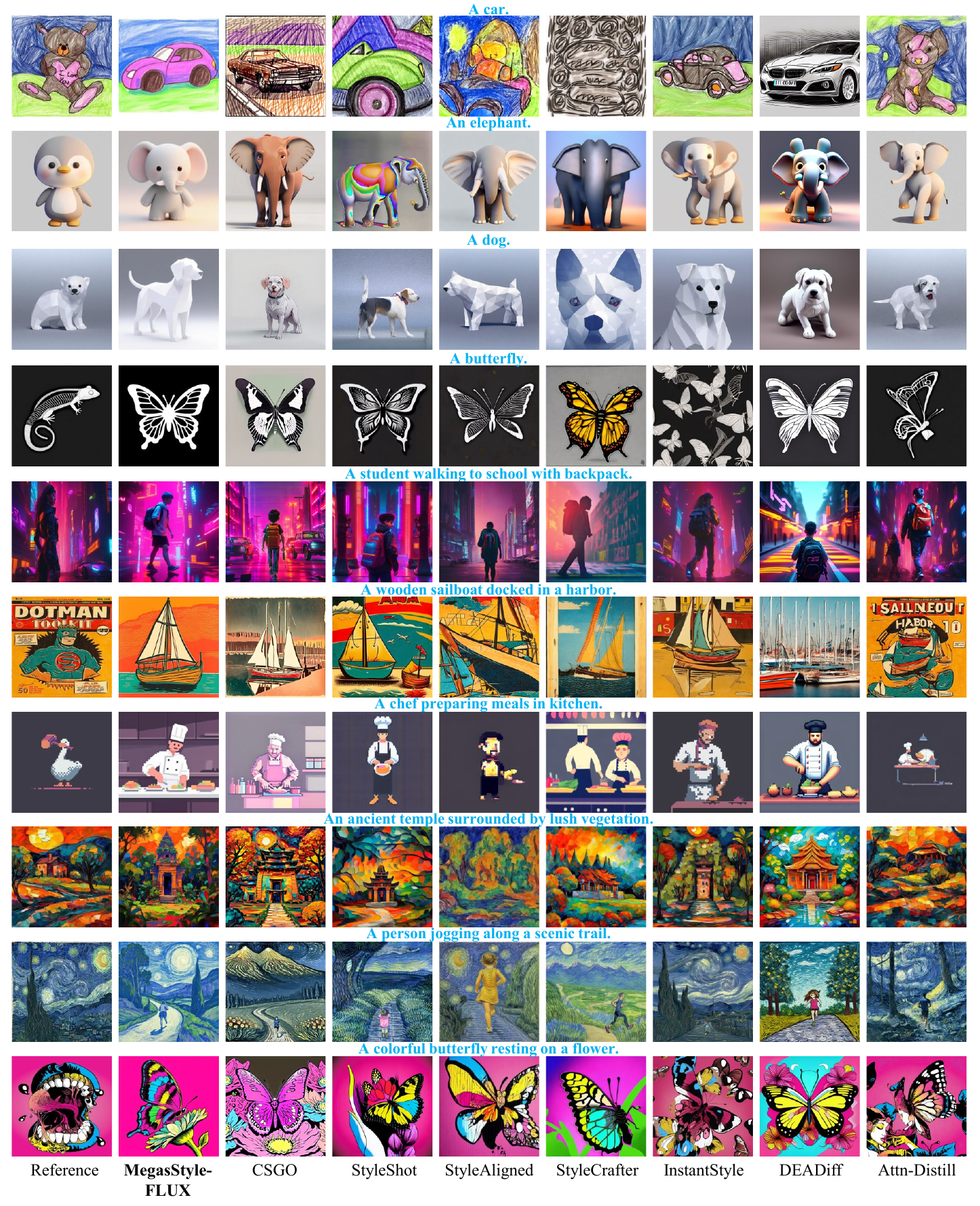}

    \caption{Additionaly qualitative comparison between MegaStyle-FLUX and SOTA style transfer methods.}
    \vspace{-4mm}
    \label{fig:comparison1}
\end{figure*}

\begin{figure*}[t]
    \centering
    \includegraphics[width=1.0\linewidth]{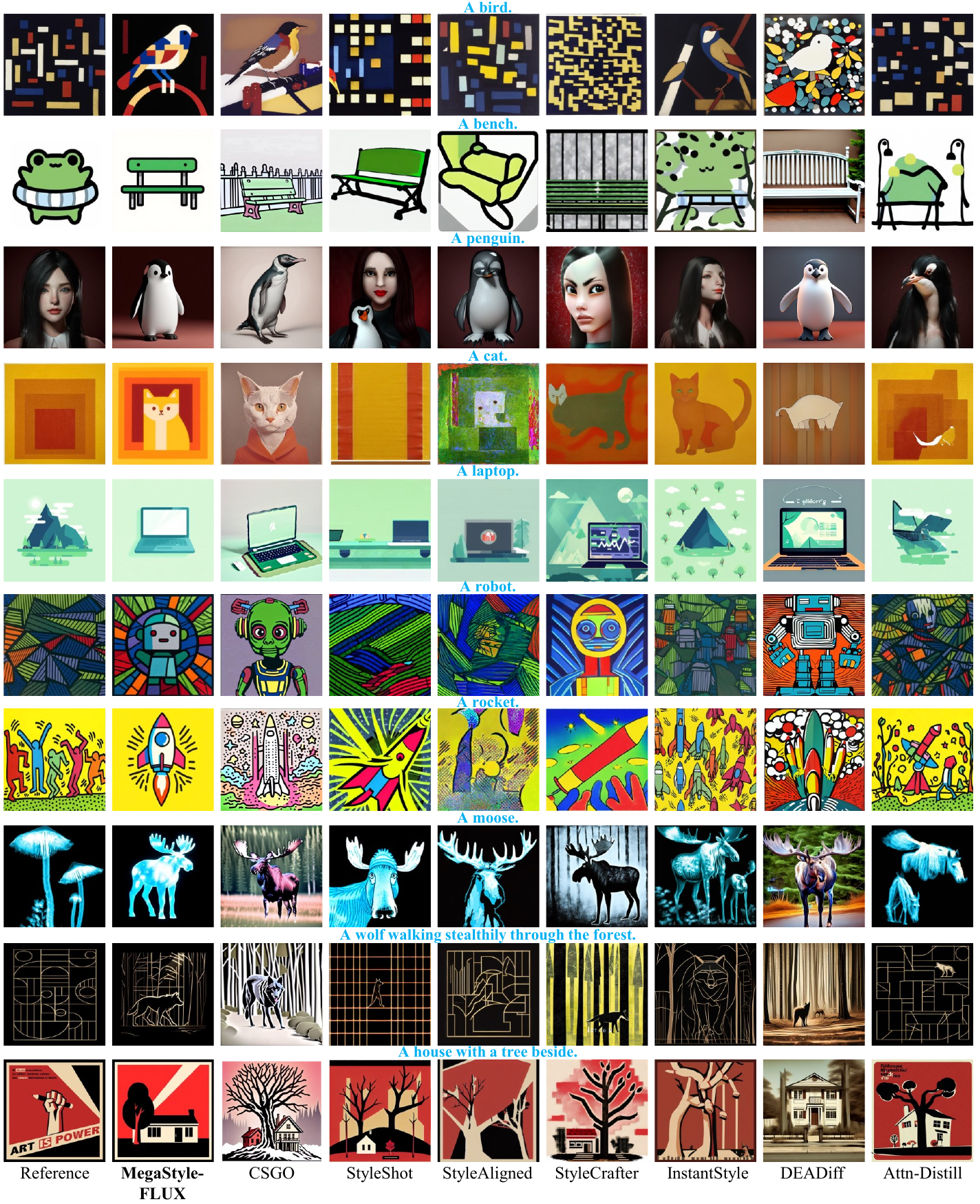}

    \caption{Additionaly qualitative comparison between MegaStyle-FLUX and SOTA style transfer methods.}
    \vspace{-4mm}
    \label{fig:comparison2}
\end{figure*}

\begin{figure*}[t]
    \centering
    \includegraphics[width=1.0\linewidth]{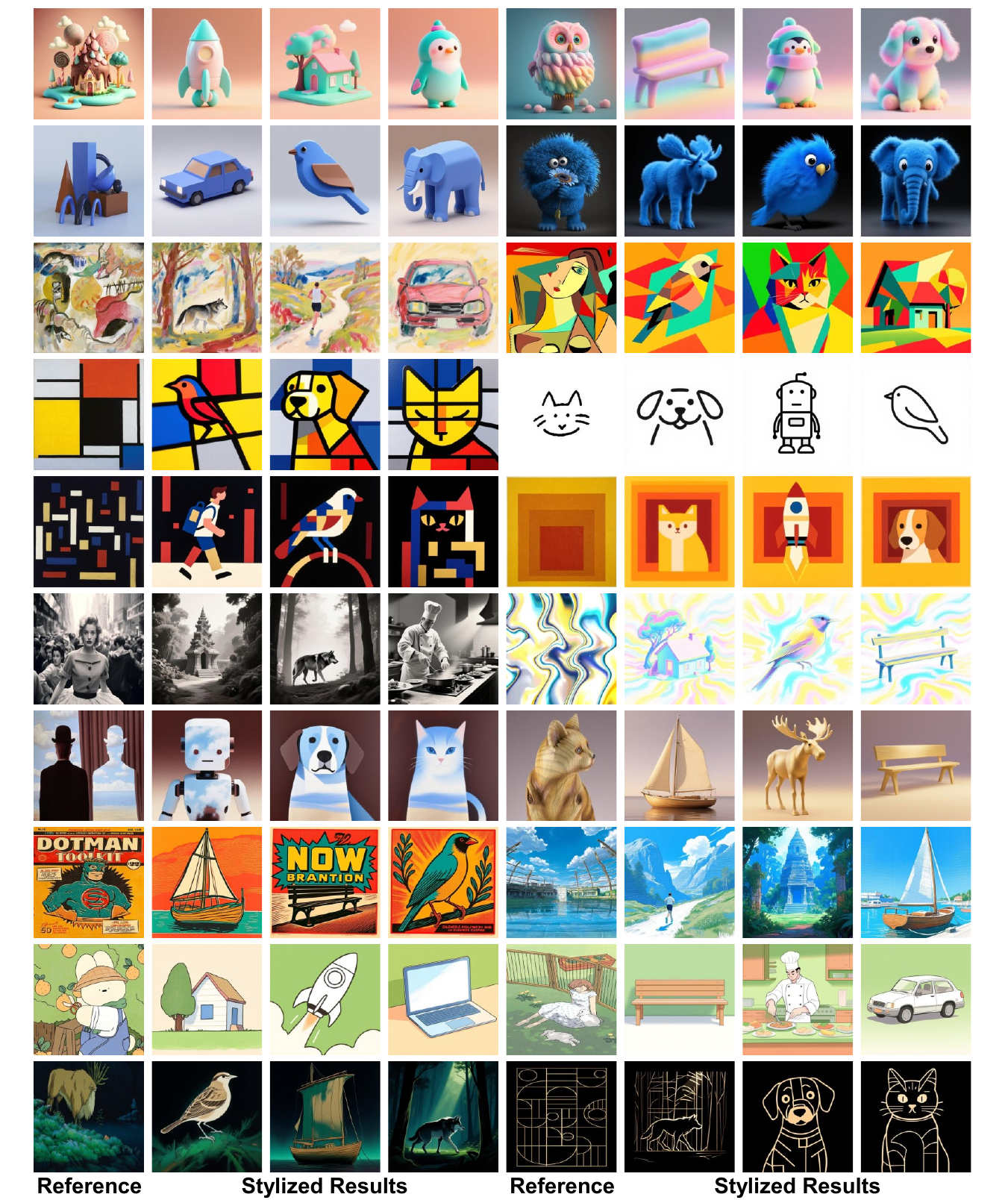}

    \caption{Stylized results of MegaStyle-FLUX.}
    \vspace{-4mm}
    \label{fig:MegaStyle1}
\end{figure*}

\begin{figure*}[t]
    \centering
    \includegraphics[width=1.0\linewidth]{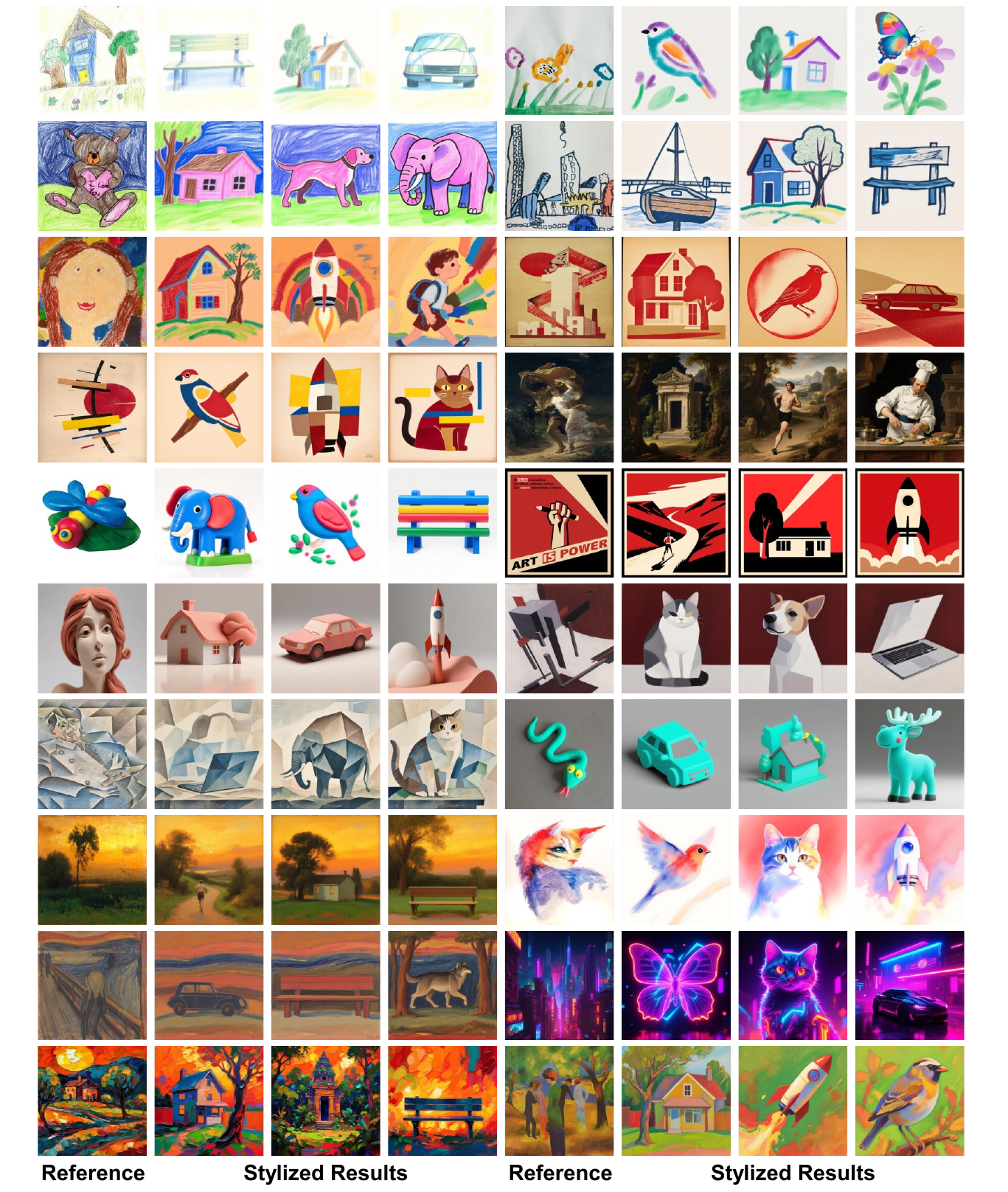}

    \caption{Stylized results of MegaStyle-FLUX.}
    \vspace{-4mm}
    \label{fig:MegaStyle2}
\end{figure*}

\begin{figure*}[t]
    \centering
    \includegraphics[width=1.0\linewidth]{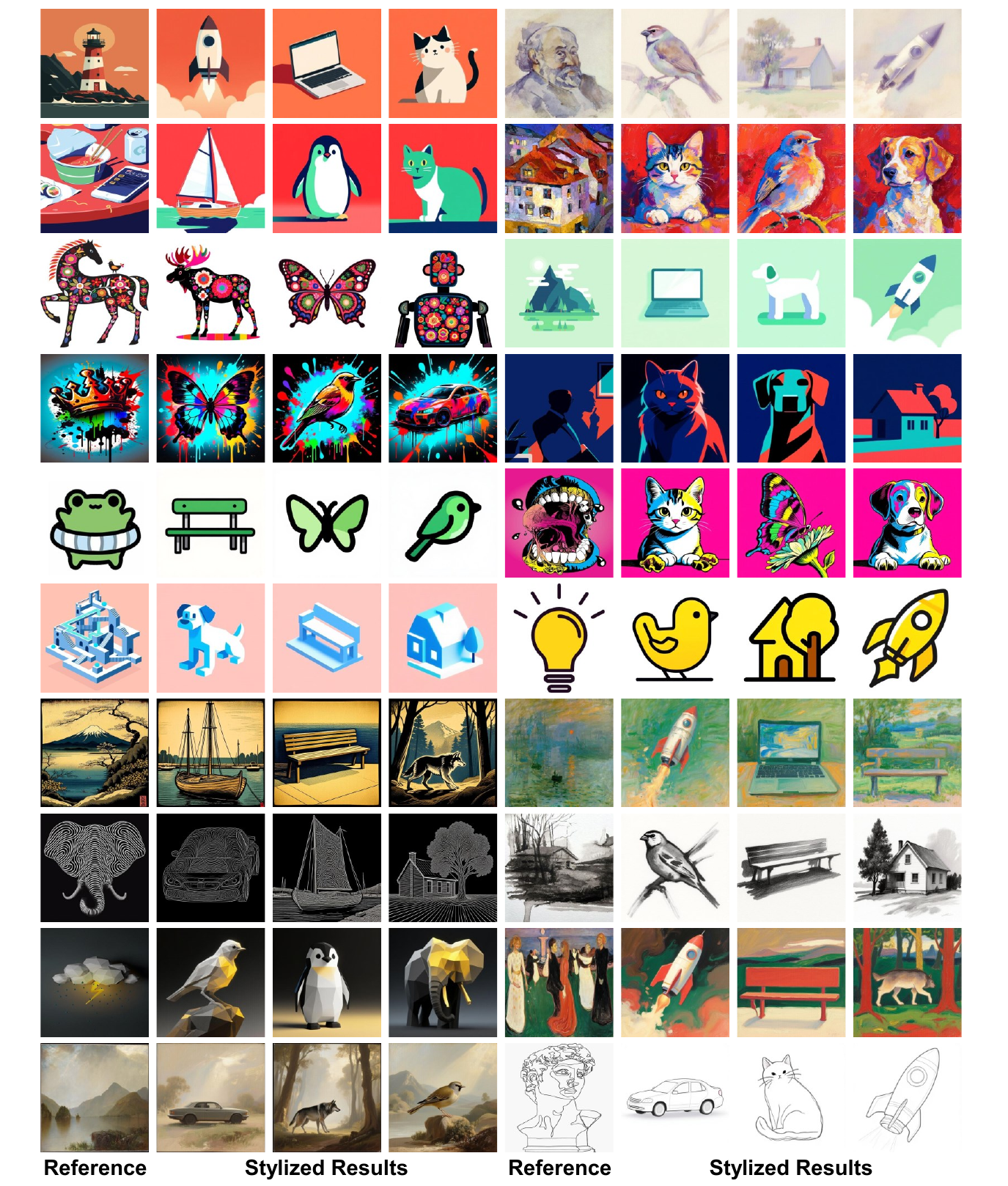}

    \caption{Stylized results of MegaStyle-FLUX.}
    \vspace{-4mm}
    \label{fig:MegaStyle3}
\end{figure*}

\begin{figure*}[t]
    \centering
    \includegraphics[width=1.0\linewidth]{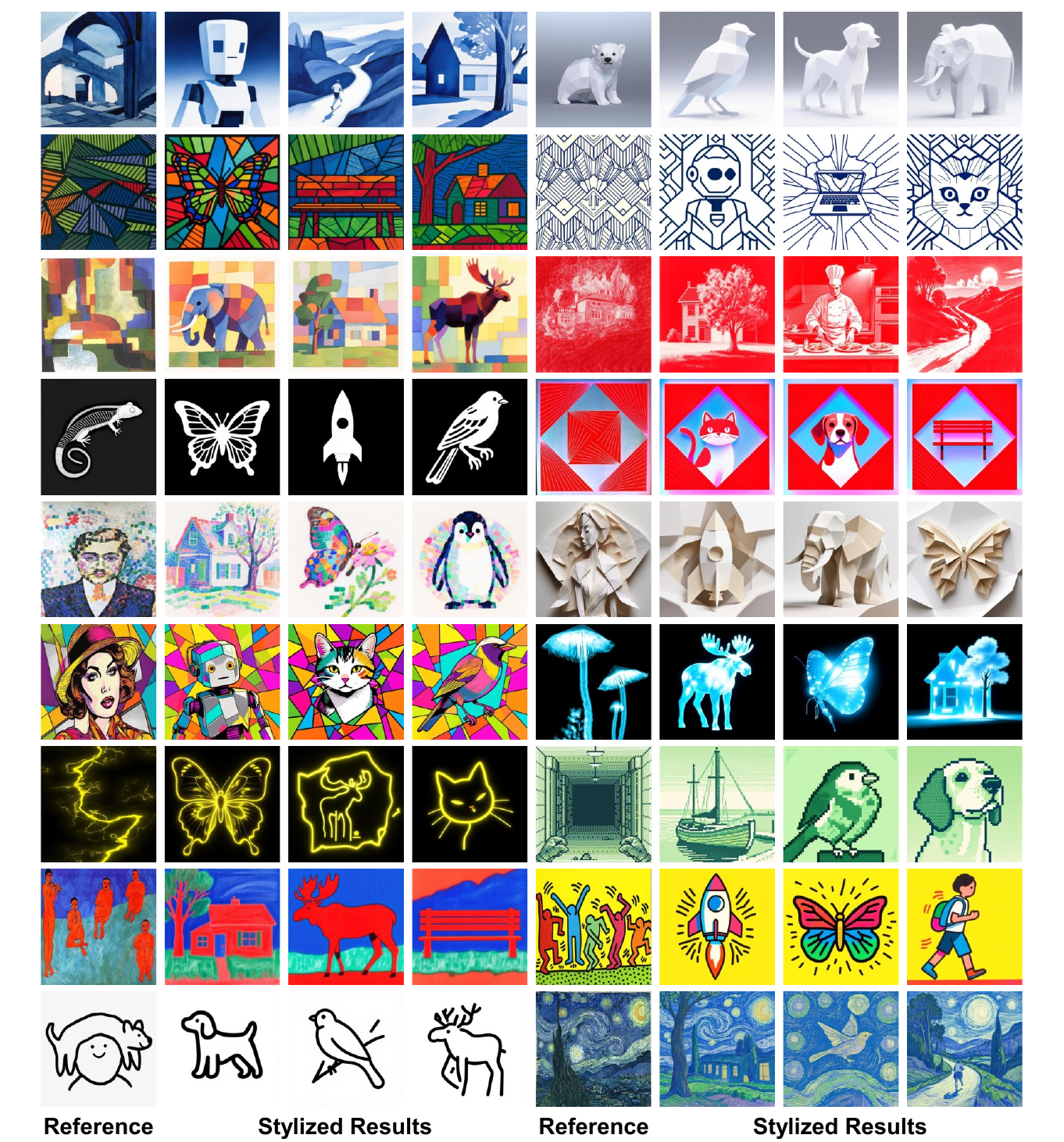}

    \caption{Stylized results of MegaStyle-FLUX.}
    \vspace{-4mm}
    \label{fig:MegaStyle4}
\end{figure*}

In this subsection, we present additional visualizations of our style dataset MegaStyle-1.4M (Figure \ref{fig:main_dataset1}, Figure \ref{fig:main_dataset2} and Figure \ref{fig:main_dataset3}), comparisons between MegaStyle-FLUX and baseline methods (Figure \ref{fig:comparison1} and \ref{fig:comparison2}) and more stylized results of MegaStyle-FLUX (Figure \ref{fig:MegaStyle1}, Figure \ref{fig:MegaStyle2}, Figure \ref{fig:MegaStyle3} and Figure \ref{fig:MegaStyle4}).

\section{Limitations}
\begin{figure}[t]
    \centering
    \includegraphics[width=1.0\linewidth]{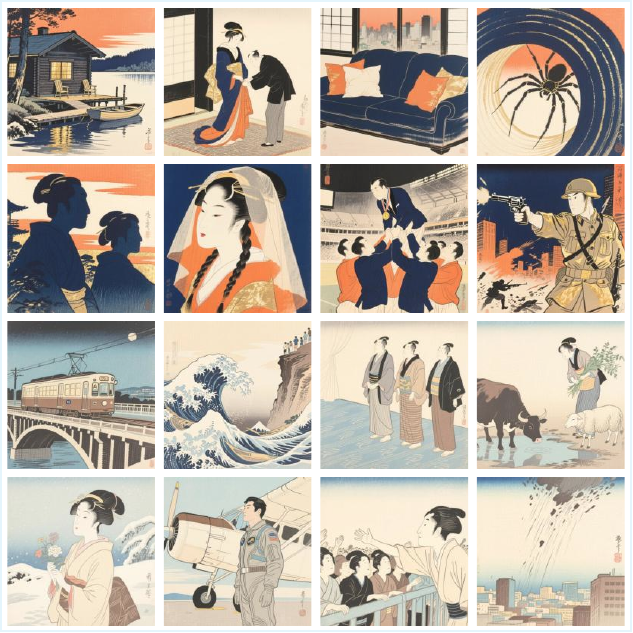}
    \vspace{-6mm}
    \caption{Visualizations of association bias in Qwen-Image. }
    \vspace{-6mm}
    \label{fig:bias}
\end{figure}

Although MegaStyle excels in constructing intra-style consistent, inter-style diverse and high-quality style dataset, some components of its data curation pipeline still have room for improvement.
For example, the generalization ability of current VLMs is limited, making it difficult for them to recognize uncommon styles.
On the other hand, Qwen-Image shows association bias toward some styles in the image generation process.
As shown in Figure \ref{fig:bias}, when the style prompt includes ``Japanese painting,'' the generated objects are often depicted as Japanese figures biased toward historical periods such as the Edo or Meiji era (e.g., kimono/yukata, traditional hairstyles, and scroll-painting–like or ancient-architecture backgrounds).
However, these limitations stem from the inherent capabilities of the models themselves. 
We will continue to closely track the latest and most powerful VLMs and T2I generation models to further improve the quality of our dataset.

\end{document}